\documentclass[runningheads]{llncs}

\usepackage{eccv}
\usepackage{amsmath}
\usepackage{algorithm}
\usepackage{algpseudocode}
\usepackage{makecell}
\usepackage{diagbox}
\usepackage[table]{xcolor}
\usepackage{array}

\usepackage{eccvabbrv}

\usepackage{graphicx}
\usepackage{booktabs}

\usepackage[accsupp]{axessibility}  
\DeclareMathOperator*{\argmax}{arg\,max}

\usepackage{hyperref}
\hypersetup{
    colorlinks=true,
    linkcolor=black,
    citecolor=black,
    urlcolor=magenta
}
\usepackage{booktabs}
\usepackage{multirow}
\usepackage{orcidlink}

\begin{document}

\title{CCFM: Collision-Constrained Flow Matching for Safety-Critical Scenario Generation} 

\titlerunning{CCFM}

\author{Ke Li\inst{1}\orcidlink{0009-0001-4958-3302} \and
Kaidi Liang\inst{1}\orcidlink{0009-0001-9129-2744} \and
Yuxin Ding\inst{2}\orcidlink{0009-0005-3629-1908} \and Debojyoti Biswas\inst{2} \orcidlink{0000-0002-8842-0207} \and Xianbiao Hu\inst{2}  \orcidlink{0000-0002-0149-1847} \and Ruwen Qin \inst{1}$^*$  \orcidlink{0000-0003-2656-8705}}

\authorrunning{K.~Li et al.}

\institute{Stony Brook University, Stony Brook, NY 11794, USA \and
Pennsylvania State University, University Park, PA 16802, USA\\
\email{\{ke.li.1, kaidi.liang, ruwen.qin\}@stonybrook.edu}\\
\email{\{ymd5170, dbb5917, xbhu\}@psu.edu}\\
$^*$ Corresponding author}

\maketitle

\begin{abstract}
  Evaluation of autonomous vehicle (AV) planners in safety-critical closed-loop simulation is essential for real-world deployment. However, generating controllable safety-critical scenarios remains challenging. Existing approaches use soft guidance that provides only probabilistic preferences and cannot guarantee the satisfaction of geometric and severity constraints associated with specific collision types. We introduce Collision-Constrained Flow Matching (CCFM), a novel framework that guarantees precise collision control through hard physical constraints. CCFM consists of three key components: (i) a heuristic collision selector that optimally identifies an adversarial agent and collision type via composite scoring; (ii) structured hard constraints that explicitly define four collision types (rear-end, side, cut-in, head-on) through contact point, heading, and severity requirements; and (iii) a collision-constrained flow matching sampler that enforces the constraints via Gauss-Newton manifold projection. CCFM achieves collision rate up to 46.4\% on nuScenes and 83.1\% on nuPlan, significantly outperforming baselines while preserving realistic driving behavior. By enabling controllable collision characteristics in safety-critical scenario generation, CCFM provides a reliable foundation for AV safety evaluation and sim-to-real crash data generation. The code and implementation details are available at \url{https://github.com/KELISBU/CCFM}.
\end{abstract}

\section{Introduction}
\label{sec:intro}

Safety-critical scenario generation is essential for evaluating the robustness of autonomous vehicles (AV) \cite{safesurvey}, as such scenarios are rare, lying in the long tail of the driving-event distribution \cite{curse} and often lead to highly consequential interactions. These events are difficult to collect at scale from naturalistic driving data due to their low frequency and high risk. Therefore, closed-loop simulation provides a critical alternative by enabling repeatable and scalable interactive AV testing. In this context, explicitly generating realistic and controllable scenarios, in which an adversarial agent is designated to challenge the ego vehicle, is essential to expose the vulnerabilities of AV planners before real-world deployment.

While recent advancements have explored safety-critical scenario generation and closed-loop simulation, significant gaps remain in balancing \textit{controllability}, precise \textit{collision constraints}, and \textit{behavior realism}. First, traditional optimization-based \cite{king} or RL-based \cite{AST} methods often induce overly aggressive maneuvers that deviate from human-like driving behavior, making it difficult to preserve realistic multi-agent interaction patterns. More recently, diffusion-based generative approaches \cite{safe-sim,ccdiff,diffscene} improve behavior realism by learning from traffic priors. However, their controllability typically relies on cost-based \textit{soft guidance}, which cannot rigorously ensure the satisfaction of spatial-geometric and severity constraints. Moreover, under dense and dynamic traffic context, pre-selecting an adversarial agent and its most feasible collision mode remains an underexplored problem.

In this work, we reformulate the safety-critical scenario generation as a collision-constrained sampling problem rather than unconstrained guided sampling or adversarial optimization. Our key idea is to preserve the learned prior of realistic traffic scenarios, while enforcing structured collision constraints that specify the desired event type. Built upon flow matching \cite{flowmatching1,flowmatching2}, which generates samples by integrating an ordinary differential equation (ODE), our framework naturally allows intermediate projection steps to correct the evolving sample toward a collision-constrained feasible set.

In summary, we propose a Collision-Constrained Flow Matching (\textbf{CCFM}) framework. The main contributions  of this paper are as follows:

\begin{itemize}
    \item Introduction of a Heuristic Collision Selector (HCS) that scores candidate agent-collision pairs based on topological reachability, geometry, and road legality, enabling dynamic identification of the most threatening yet physically viable interaction.
\item Formulation of a comprehensive collision-type-specific constraint set that mathematically formulates the physical constraints governing the contact point, relative heading, and severity metrics for different collision types in adversarial trajectory generation.
\item Development of a Gauss-Newton manifold projection method that projects intermediate action sequences onto the feasible set while preserving plausible driving behavior.

\end{itemize}

Extensive experiments on nuScenes \cite{nuscenes} and nuPlan \cite{nuplan} closed-loop simulation illustrate that CCFM generates more safety-critical scenarios than prior methods, while achieving interpretable collision-type controllability, diverse collision outcomes, and competitive realism.
\section{Related Work}
\label{sec:Related Work}
\subsection{Generative Models for Traffic Scenario Generation}

Early data-driven approaches established the foundation for multimodal interactive trajectory generation using diverse paradigms, including conditional variational autoencoders (CVAEs) \cite{Trajectron}, generative adversarial networks (GANs) \cite{social_gan}, energy-based models \cite{Energy-Based-Model}, and  autoregressive transformers \cite{BehaviorGPT,trafficgen}. To achieve better distributional expressivity and controllability, diffusion models \cite{diffusion1,diffusin2} and Flow Matching \cite{flowmatching1,flowmatching2,flowmatching3} methods have recently emerged as the dominant baseline. By framing generation as a progressive denoising process, frameworks such as Diffuser \cite{diffuser} unify generative priors with planning objectives through classifier-guided sampling. However, Flow Matching provides a more efficient continuous-time formulation for distribution transport. For instance, Flow Planner \cite{flowplanner} and GoalFlow \cite{goalflow} successfully leverage it to model interactive multi-agent behaviors and goal-driven multimodal trajectories, demonstrating significant inference speedups and robust performance on large-scale planning benchmarks.

\subsection{Safety-critical Scenario Generation with Controllability}

Rigorous validation of autonomous vehicles necessitates closed-loop simulation to generate rare, but safety-critical scenarios. Existing generation paradigms can be broadly grouped into (i) optimization-based methods, such as AdvSim \cite{advsim}, ReGentS \cite{ReGentS}, and KING \cite{king}, which iteratively perturb trajectories toward risk; (ii) hybrid approaches combining learned priors with optimization and rule-based rollouts, like STRIVE \cite{strive} and SLEDGE \cite{SLEDGE}. More recently, frameworks such as DiffScene \cite{diffscene} and SAFE-SIM \cite{safe-sim} exemplify the diffusion-based paradigm, employing guided cost functions to steer realistic multi-agent distributions toward collision-prone outcomes during closed-loop rollouts. Recent studies further improve controllability from different perspectives, including closed-loop adversarial resampling~\cite{CAT}, feasibility-guided adversarial policy learning~\cite{FREA}, inverse motion prediction for inserted adversarial agents~\cite{AdvBMT}, and preference-aligned steerable
scenario generation~\cite{SAGE}. To further enhance high-level controllability, recent studies integrate large language models (LLMs). For instance, LD-Scene \cite{Ld-scene} translates natural language intents into executable adversarial loss functions for latent diffusion, whereas ChatScene \cite{chatscene} acts as an autonomous agent that bridges unstructured text into structured configurations of CARLA simulator.

Diffusion models steer generative distributions toward risk via soft guidance \cite{safe-sim,ccdiff,Ld-scene}, rather than enforcing hard constraints, thereby limiting controllability. To overcome this limitation, recent works such as PCFM \cite{pcfm} and HardFlow \cite{hardflow} enforce hard constraints at inference time by projecting intermediate flow states onto the constraint manifold without retraining. However, both methods have been developed and evaluated exclusively in the context of physics-governed PDE systems. Whether such hard-constraint enforcement transfers to interactive scenario generation remains an open question, since traffic constraints are dynamics-mediated and defined relative to an evolving ego trajectory.

\section{Methodology}
\label{sec:Methodology}
\begin{figure}[htb]
    \centering
    \includegraphics[width=1.0\linewidth]{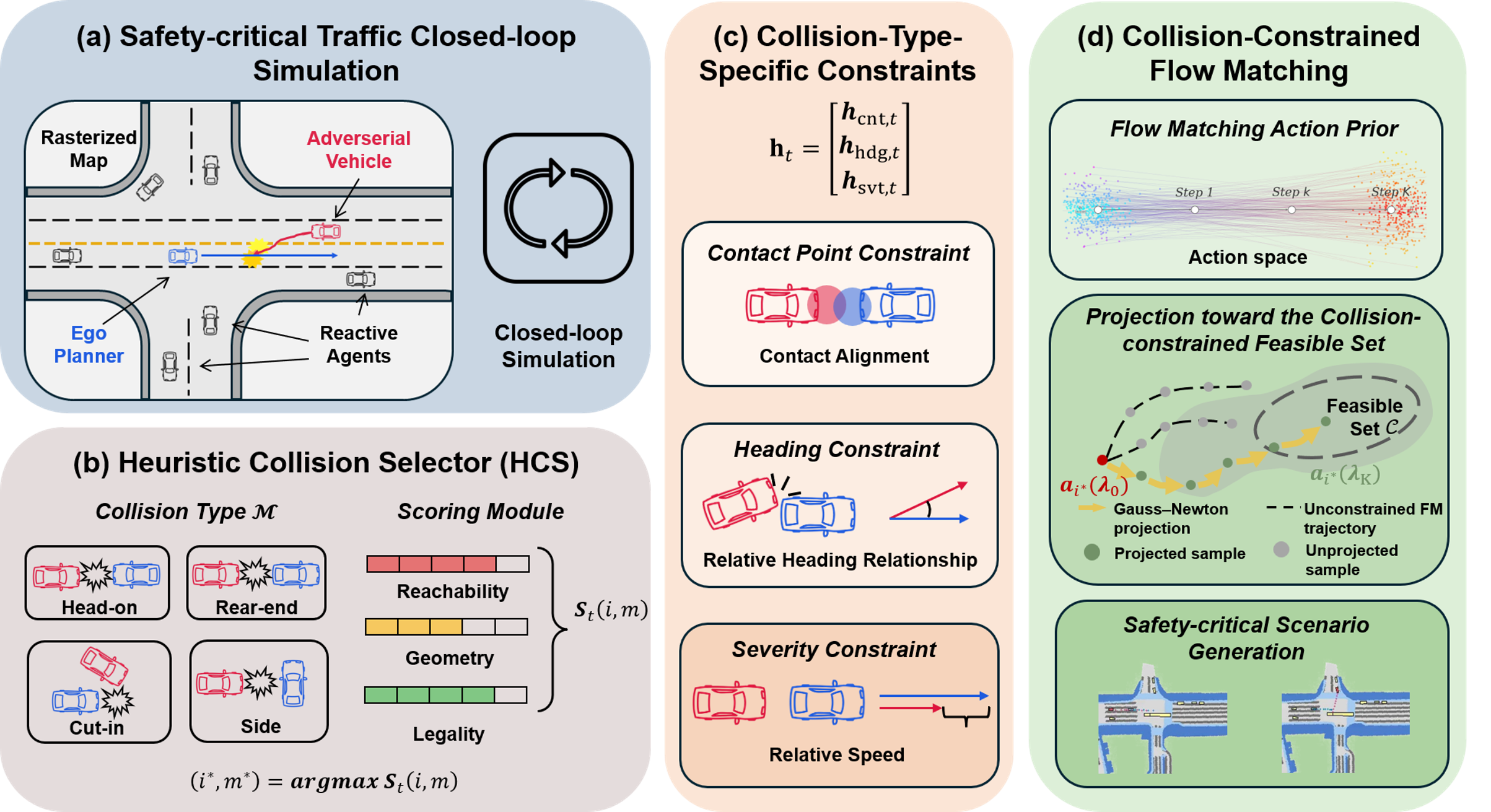}
    \caption{Overview of the proposed safety-critical scenario generation by CCFM.}
    \label{fig:CCFM}
\end{figure}

We propose Collision-Constrained Flow Matching (CCFM), a framework for safety-critical scenario generation, illustrated in Fig. \ref{fig:CCFM}. CCFM is formulated within a closed-loop simulation environment. The Heuristic Collision Selector (HCS) dynamically identifies the most threatening yet physically viable collision type and the associated adversarial vehicle. For each collision type, a set of collision-type-specific constraints are defined, resulting in the collision-constrained Flow Matching problem. CCFM is solved via a damped Gauss-Newton projection for generating the desired safety-critical scenario. Details of the methodology are presented as follows.

\subsection{Safety-critical Traffic Generation in Closed-loop Simulation}
\label{Sec3.1 problem formulation}

In the closed-loop simulation, we consider a traffic scene with $N$ vehicles: an \emph{ego vehicle} ($i=\mathrm{ego}$) and $N-1$ non-ego vehicles. One non-ego vehicle is selected as the \emph{adversarial agent} and tasked with colliding with the ego vehicle, while the others are \emph{reactive agents} that respond to traffic to preserve realism. Our goal is to generate safety-critical scenarios that satisfy controllable conditions for any specific collision type.




$\mathbf{s}_{i,t} = [x_{i,t}, y_{i,t}, \vartheta_{i,t}, v_{i,t}]$ denotes the state of vehicle $i$ at timestep $t$, $\forall\, i \in \{1,\ldots,N-1\}\cup\{\mathrm{ego}\}$.  $\mathbf{p}_{i,t}=[x_{i,t},y_{i,t}]$ represents vehicle $i$'s 2D position, while $\vartheta_{i,t}$ and $v_{i,t}$ denote its heading angle and linear velocity, respectively. At timestep $t$,  the action of vehicle $i$, $\mathbf{a}_{i,t} = [\dot{v}_{i,t}, \dot{\vartheta}_{i,t}]$, consists of its acceleration and yaw rate. The selection of this action is conditioned on the vehicle's context, $\mathbf{c}_{i,t}$, defined in the supplementary material. The state evolution is governed by a one-step deterministic unicycle dynamics model $f$:
\begin{equation}
\mathbf{s}_{i,t+1} = f(\mathbf{s}_{i,t}, \mathbf{a}_{i,t}).
\end{equation}

The ego vehicle drives autonomously, with its future trajectory $\boldsymbol{\tau}_{\mathrm{ego},t:t+T}$ controlled by an AV planner $\pi$:
\begin{equation}
\boldsymbol{\tau}_{\mathrm{ego},t:t+T} = \pi(\mathbf{c}_{\mathrm{ego},t}).
\end{equation}

Non-ego vehicles are governed by a generative prior $p_\theta(\cdot \mid \mathbf{c}_t)$ learned via Flow Matching, which captures the distribution of realistic driving trajectories. We generate the action sequence for each non-ego vehicle, $\mathbf{a}_{i,t:t+T-1}$, from this learned prior:
\begin{equation}
\mathbf{a}_{i,t:t+T-1} \sim p_\theta(\cdot \mid \mathbf{c}_{i,t}), 
\label{eq:action sequence}
\end{equation}
and obtain the corresponding trajectory, $\boldsymbol{\tau}_{i,t:t+T}$, using a forward $T$-step rollout function $\mathcal{F}$:
\begin{equation}
    \boldsymbol{\tau}_{i,t:t+T} = \mathcal{F}(\mathbf{s}_{i,t},\mathbf{a}_{i,t:t+T-1}).
    \label{eq:F}
\end{equation}
The generation of trajectories for reactive agents is \emph{unconstrained}, whereas  the sampled action sequence for the adversarial agent must yield a trajectory satisfying the hard physical constraints for a given collision type, to be introduced in Sec. \ref{Sec3.3}.

In a receding-horizon closed-loop simulation, all vehicles update their contexts from the latest states, re-plan their horizon-$T$ action sequences, execute only the first action, and repeat.

\subsection{Heuristic Collision Selector}
\label{Sec3.2 Heuristic Collision Selector}

At each timestep $t$, the selection of the most plausible safety-critical scenario to generate is formulated as an optimization problem that jointly identifies the adversarial agent and its specific collision type with the ego vehicle. Directly optimizing over all candidate agents and collision types when solving the constrained Flow Matching problem for trajectory generation is computationally inefficient and often physically implausible. Therefore, we decouple this selection process from the constrained Flow Matching formulation by introducing a Heuristic Collision Selector (HCS) to identify an optimal pair of adversarial agent and collision type, $(i^\ast, m^\ast)$, that maximizes the collision score function:
\begin{equation}
S_t(i,m) = w_{\mathrm{rch}} S_{\mathrm{rch},t}(i) + w_{\mathrm{geo}} S_{\mathrm{geo},t}{(i, m)} + w_{\mathrm{lgt}} S_{\mathrm{lgt},t}{(i, m)},
\label{eq:collision score}
\end{equation}
where $i \in \{1, \dots, N-1\}$ and $m \in \mathcal{M} = \{\text{Rear-End}, \text{Side}, \text{Cut-In}, \text{Head-On}\}$. The collision score in Eq.~\eqref{eq:collision score} is defined  as a weighted sum of the reachability score ($S_{\mathrm{rch},t}$), the geometry score ($S_{\mathrm{geo},t}$), and the legality score ($S_{\mathrm{lgt},t}$), with weights $w_{(\cdot)}$ summing to one. These similarity-based scores are outlined below, and their modeling details are provided in the supplementary material.

The reachability score for agent $i$ at timestep $t$ is calculated from its position relative to the ego vehicle:
\begin{equation}
    S_{\mathrm{rch},t}(i) = \psi_\mathrm{rch}(\mathbf{p}_{i,t}, \mathbf{p}_{\text{ego},t}),
\end{equation}
where $\psi_{\mathrm{rch}}(\cdot)$ measures the proximity-based similarity between the two vehicles, reflecting the higher collision likelihood of nearby agents.

The geometry score for agent $i$ under collision type $m$ is computed by comparing the agent's relative pose with respect to the ego vehicle at time $t$, $\mathbf{pos}_{i,t}$, against the typical pose associated with that collision type, $\mathbf{pos}_m$:
\begin{equation}
S_{\mathrm{geo},t}(i,m) = \psi_{\mathrm{geo}}(\mathbf{pos}_{i,t}, \mathbf{pos}_m),
\end{equation}
where $\mathbf{pos}_{i,t}$ is derived from the positions and heading angles of the two vehicles, and $\psi_{\mathrm{geo}}(\cdot)$ measures the similarity between $\mathbf{pos}_{i,t}$ and the reference pose $\mathbf{pos}_m$.

The legality score is computed by comparing the topological relationship between agent $i$ and the ego vehicle at timestep $t$, $\mathbf{tpl}_{i,t}$, against the typical relationship for collision type $m$, $\mathbf{tpl}_m$:
\begin{equation}
S_{\mathrm{lgt},t}(i,m) = \psi_{\mathrm{lgt}}(\mathbf{tpl}_{i,t}, \mathbf{tpl}_m),
\end{equation}
where $\mathbf{tpl}_{i,t}$ is determined based on the contexts, positions, and heading angles of the two vehicles. $\psi_{\mathrm{lgt}}(\cdot)$ measures the similarity between $\mathbf{tpl}_{i,t}$ and $\mathbf{tpl}_m$, ensuring that the generated adversarial behavior remains map-compliant and realistic.

We note that HCS is implemented as a plug-in function that can be overridden by users’ manual selections, providing flexibility to generate user-specified scenarios.

\subsection{Collision-Type-Specific Constraints}
\label{Sec3.3}

At timestep $t$, after HCS identifies the adversarial agent $i^\ast$ and the collision type $m^\ast$ for the generation of safety-critical scenario,  a set of constraints is imposed to ensure that the adversarial agent's trajectory will lead to the desired collision type at the future timestep $t+T_\mathrm{col}$. Here, $T_\mathrm{col}$ is the target time-to-collision, with details of its selection provided in the supplementary material.

We formulate these constraints through residual measures:
\begin{equation}
\mathbf{h}_t=
\begin{bmatrix}
    h_{\mathrm{cnt},t}\\
    h_{\mathrm{hdg},t}\\    
    h_{\mathrm{svt},t}
    \end{bmatrix},
    \label{eq:h}
\end{equation}
where $h_{\mathrm{cnt},t}$, $h_{\mathrm{hdg},t}$, and $h_{\mathrm{svt},t}$ denote the residual functions related to contact-point (cnt), relative-heading (hdg), and impact severity (svt) constraints, respectively:
\begin{equation}
    h_{j,t} = \phi_j(l_{j,t}, \bar{l}_j),\quad \text{for } j \in \{\mathrm{cnt}, \mathrm{hdg}, \mathrm{svt}\}.
    \label{eq:hjt}
\end{equation}
In eq. (\ref{eq:hjt}), $l_{j,t}$ denotes the $j$-measure computed from the predicted states of the adversarial agent and the ego vehicle at the target time of impact $t+T_\mathrm{col}$, and $\bar{l}_j$ represents the corresponding target value for collision type $m^\ast$. The function $\phi_j(\cdot)$ maps $l_{j,t}$ and $\bar{l}_j$ to a dissimilarity score, enabling the related constraints to be imposed in action sequence generation. Details of the derivations for $l_{j,t}$, $\bar{l}_j$, and $\phi_j$ for all $j$ are provided in the supplementary material.

\subsection{Collision-Constrained Flow Matching}
\label{Sec3.4 Collision-Constrained Flow Matching}

\subsubsection{Unconstrained Action Sequence Sampling.}
Following the conditional Flow Matching method \cite{flowmatching2}, a velocity field $\mathbf{v}_\theta(\lambda, \mathbf{a}(\lambda), \mathbf{c})$ is learned, where $\lambda\in[0,1]$ is the flow time variable. At inference time, to sample an action sequence for agent $i$, $\mathbf{a}_{i,t:t+T-1}$, we draw an initial action sequence from a standard Gaussian distribution: $\mathbf{a}_i(\lambda_0) \sim \mathcal{N}(\mathbf{0}, \mathbf{I})$, and then transport it to the learned conditional action prior $p_\theta(\cdot|\mathbf{c}_{i,t})$ by solving the ordinary differential equation (ODE):
\begin{equation}
  d \mathbf{a}_i(\lambda)/d\lambda 
  = \mathbf{v}_\theta(\lambda, \mathbf{a}_i(\lambda), \mathbf{c}_{i,t}),
  \label{eq:ODE}
\end{equation}
from $\lambda{=}0$ to $\lambda{=}1$. The terminal point $\mathbf{a}_i(1)$ is the planned action sequence $\mathbf{a}_{i,t:t+T-1}$ in Eq. (\ref{eq:action sequence}), which is subsequently propagated 
through the dynamics model $\mathcal{F}$ in Eq. (\ref{eq:F}) to produce the future state trajectory 
$\boldsymbol{\tau}_{i,t:t+T}$.

We numerically integrate the unconstrained ODE in Eq. (\ref{eq:ODE}) using a discrete Euler solver with $K$ steps. The intermediate sample is updated at each integration step $k=0,1,\dots, K-1$: 
\begin{equation}
\mathbf{a}_i(\lambda_{k+1})
=
\mathbf{a}_i(\lambda_k)
+
\Delta\lambda \cdot
\mathbf{v}_\theta(\lambda_k,\mathbf{a}_i(\lambda_k),\mathbf{c}_{i,t}),
\label{eq:euler_step}
\end{equation}
where $\Delta\lambda = 1/K$ and $\lambda_k = k\Delta\lambda$. However, this unconstrained integration alone does not guarantee that the adversarial trajectory satisfies the desired collision conditions.

\subsubsection{Manifold Projection in Flow Matching.}
\label{Sec.3.4.2 Manifold Projection in Flow Matching}
With the set of three residual functions $\mathbf{h}_t(\cdot)$ defined in Sec.\ref{Sec3.3}, CCFM enforces collision conditions when sampling adversarial action sequences. Following the step-wise projection strategy of PCFM \cite{pcfm}, we project intermediate action sequences onto the collision-constrained feasible set at every ODE step, and propagate the projected sequence back to the current flow time via an optimal-transport (OT) reverse update.

\paragraph{Collision-constrained Feasible Set.}
At each integration step, the intermediate action sequence is projected onto the collision-constrained feasible set:
\begin{equation}
\mathcal{C}=\left\{
\mathbf{a}_{i^*,t:t+T-1}\;\middle|\;
\mathbf{h}_t=\mathbf{0}
\right\},
\label{eq:feasible_set}
\end{equation}

where $\mathbf{h}_t$ is the set of  residual functions defined in Eq.(\ref{eq:h}). We denote the corresponding projection operator by $\Pi_{\mathcal{C}}:\mathbb{R}^{T\times 2}\!\to\!\mathcal{C}$, which maps an action sequence to the closest sequence satisfying $\mathbf{h}_t 
=\mathbf{0}$ at the dynamic time-to-collision $T_\mathrm{col}$. Here, $T_\mathrm{col}$ is updated once per closed-loop re-planning step. In practice we implement $\Pi_{\mathcal{C}}$ as a damped Gauss-Newton (GN) projection with Jacobians obtained by automatic differentiation through the forward-dynamics rollout $\mathcal{F}$. Full algorithmic details of the dynamic time-to-collision and GN projection are provided in the supplemental material.

\paragraph{OT Reverse Update.}
Applying $\Pi_{\mathcal{C}}$ directly at the current flow time $\lambda_{k+1}$ produces a large discontinuity when $\lambda_{k+1}$ is small, since the unprojected sample is still far from the data manifold. Similar to PCFM \cite{pcfm}, we blend the projected sequence with the initial noise $\mathbf{a}_{i^*}(\lambda_0)$ along the optimal-transport interpolant:
\begin{equation}
\mathbf{a}_{i^*}(\lambda_{k+1})
\;=\;\lambda_{k+1}\,\Pi_{\mathcal{C}}\!\left(\hat{\mathbf{a}}_{i^*}(\lambda_{k+1})\right)
+(1-\lambda_{k+1})\,\mathbf{a}_{i^*}(\lambda_0),
\label{eq:ot_reverse}
\end{equation}
where $\hat{\mathbf{a}}_{i^*}(\lambda_{k+1})=\mathbf{a}_{i^*}(\lambda_k)+\Delta\lambda\,\mathbf{v}_\theta(\lambda_k,\mathbf{a}_{i^*}(\lambda_k),\mathbf{c}_{i^*,t})$ is the unprojected Euler step. Eq. (\ref{eq:ot_reverse}) recovers the unprojected sample as $\lambda_{k+1}\!\to\!0$ and the fully-projected sample at $\lambda_{k+1}\!=\!1$, so the constraint is satisfied exactly at the terminal step while early-step iterates remain close to the learned flow. The sampling procedure is summarized in Algorithm \ref{alg:ccfm}.

\begin{algorithm}[tb]
\caption{Collision-Constrained Flow Matching (CCFM) Sampling}
\label{alg:ccfm}
\begin{algorithmic}[1]
\State \textbf{Input:} Adversarial agent's context $\mathbf{c}_{i^\ast,t}$ and state $\mathbf{s}_{i^\ast,t}$, ego trajectory $\boldsymbol{\tau}_{\text{ego},t:t+T}$, target collision type $m^*$, ODE integration steps $K$
\State \textbf{Output:} Adversarial action sequence $\mathbf{a}_{i^\ast,t:t+T-1}$
\State Sample $\mathbf{a}_{i^*}(\lambda_0) \sim \mathcal{N}(\mathbf{0},\mathbf{I})$;\quad $\Delta\lambda \gets 1/K$
\State Compute $T_{\mathrm{col}}$ from current ego–adversary kinematics \Comment{ See supplementary}
\For{$k = 0, 1, \ldots, K-1$}
    \State $\lambda_k \gets k\cdot\Delta\lambda$
    \State $\mathbf{v} \gets \mathbf{v}_\theta(\lambda_k,\mathbf{a}_{i^\ast}(\lambda_k),\mathbf{c}_{i^*,t})$ \Comment{Predict flow velocity}
    \State $\hat{\mathbf{a}}_{i^*}(\lambda_{k+1}) \gets \mathbf{a}_{i^*}(\lambda_k) + \Delta\lambda\cdot\mathbf{v}$ \Comment{Euler step}
    \State $\mathbf{a}_{\mathrm{proj}} \gets \Pi_{\mathcal{C}}(\hat{\mathbf{a}}_{i^*}(\lambda_{k+1}))$ \Comment{Damped GN projection; see supplementary}
    \State $\mathbf{a}_{i^\ast}(\lambda_{k+1}) \gets \lambda_{k+1}\,\mathbf{a}_{\mathrm{proj}} + (1-\lambda_{k+1})\,\mathbf{a}_{i^*}(\lambda_0)$ \Comment{OT reverse update}
\EndFor
\State \Return $\mathbf{a}_{i^\ast}(\lambda_K)$
\end{algorithmic}
\end{algorithm}

\section{Experimental Setup}
\label{Sec.4 Experiments and Results}

\subsection{Datasets}
\label{sec 4.1 dataset}
We conduct extensive experiments on the real-world driving dataset nuScenes \cite{nuscenes}, and nuPlan \cite{nuplan}. Our CCFM model is trained on the nuScenes train split and evaluated in the closed-loop simulation on the nuScenes validation split and nuPlan mini validation split. In addition to the standard planning setting with an 8-second horizon, we further perform the closed-loop simulation over the duration of 20 seconds.

\subsection{Implementation Details}
\label{sec: 4.2 implementation details}
\textbf{Closed-loop Simulation Setup.}
Our closed-loop traffic simulation  is built upon TBSim \cite{tbsim} with a re-planning frequency of 2 Hz. We consider three planners for the ego vehicle: (i) a rule-based planner operating on the lane graph following STRIVE \cite{strive}, (ii) PDM \cite{pcm-closed}, a deterministic closed-loop planner, and (iii) the Intelligent Driver Model (IDM) \cite{IDM}. To evaluate the effectiveness of our approach, we compare CCFM against three baselines for closed-loop simulation: STRIVE \cite{strive}, CCDiff \cite{ccdiff}, and SAFE-SIM \cite{safe-sim}.

\textbf{Flow Matching Model.}
Following previous studies \cite{safe-sim,ctg}, we encode the rasterized map and contextual trajectories using a ResNet \cite{resnet} backbone, and adopt a U-Net architecture with several 1D convolutional blocks to model temporal consistency. During training, we use 10 historical frames (i.e., $T_\mathrm{hist}$=10) as context and predict the 32 future frames (i.e., $T$=32). Additional details about training and inferring the Flow Matching model are provided in the supplementary material.

\subsection{Evaluation Metrics}
\label{sec: 4.3 evaluation details}
To generate controllable and diverse safety-critical scenarios with realistic behaviors, we validate the proposed CCFM method on three dimensions: scenario criticality, behavior realism, and scenario controllability and diversity.

\textbf{Scenario Criticality.}
We quantify the criticality of generated scenarios using (i) the Collision Rate (CR) and (ii) the collision severity.
Specifically, CR is defined as the percentage of simulated scenes in which at least one collision involving the ego vehicle and the adversarial agent. For collision severity, we compute the value of relative speed (MS) between the ego and the adversarial vehicles at the collision time, and report the mean value over all collision events.

\textbf{Behavioral Realism.}
Adopting the approach from SAFE-SIM \cite{safe-sim} and CTG \cite{ctg}, we evaluate realism (RM) by computing the Wasserstein distance between ground truth and generated scene on normalized histograms of the driving longitudinal acceleration, lateral acceleration, and jerk. 
In addition, the Off-Road (OR) rate reports the ratio of reactive agents driving off-road across the scene level.

\textbf{Scenario Controllability and Diversity.}
We assess controllability by measuring the ratio of generated safety-critical events that match the desired event specification. Specifically, we compute the collision Type-Match (TM) rate between the generated event and the target type selected from HCS. To quantify diversity among generated safety-critical outcomes, we measure the distribution of collision contact regions on the ego vehicle at impact, reported as the Front, Rear, and Side rates by entropy score (EN).
Meanwhile, we report the standard deviation of (i) the relative speed (SD) and (ii) the relative heading (HD) at first impact.

We also report the wall-clock runtime of a full closed-loop simulation (TIME), and provide detailed formulations of all metrics and composite scores, including the Criticality Score (CS), Diversity Score (DS), Realism Score (RS), and the overall Composite Weighted Score (CWS), in the supplementary material.
\section{Evaluation and Analysis}
\label{Sec.5 evaluation and analysis}

In this section, we evaluate the proposed CCFM framework for controllable safety-critical scenario generation in closed-loop traffic simulation on nuScenes and nuPlan. Unless otherwise specified, all experiments are conducted on nuScenes. The results show that our method can generate safety-critical scenarios with explicit control over desired collision types while preserving physical plausibility.

\begin{table}[t]
\caption{Closed-loop simulation results under different rollout horizons. CR is the collision rate for adversarial agent and OR is the off-road rate for reactive agents. $^\dagger$ denotes relative composite scores normalized within this comparison table.}
\centering
\setlength{\tabcolsep}{2.5pt}
\begin{tabular}{l|l| c c |c  >{\columncolor{gray!12}}c |c c  >{\columncolor{gray!12}}c |c}
\toprule
\textbf{Dataset}&
\textbf{Model} 
& \makecell{\textbf{CR}\\ (\%) $\uparrow$} 
& \makecell{\textbf{MS}\\ ($m/s$) $\uparrow$} 
& \makecell{\textbf{TM}\\(\%) $\uparrow$} 
& \makecell{\textbf{CS}$^\dagger$\\ $\uparrow$}
& \makecell{\textbf{OR}\\ (\%) $\downarrow$}  
& \makecell{\textbf{RM}\\ $\downarrow$} 
& \makecell{\textbf{RS}$^\dagger$\\ $\uparrow$}
& \makecell{\textbf{Time}\\(s) $\downarrow$} \\
\midrule
\multicolumn{10}{l}{\textbf{Horizon = 80}} \\
\midrule
\multirow{4}{*}{nuScenes\cite{nuscenes}}
& CCDiff \cite{ccdiff}     & 4.0 & 0.3 & -- &0.10  & \textbf{4.4} & \textbf{0.49} & \textbf{1.00} & 180.5 \\
& STRIVE \cite{strive}     & 20.8 & 1.9 & -- & 0.56 & 5.8  & 0.83 & 0.67 & 466.3 \\
& SAFE-SIM \cite{safe-sim} & 25.8 & 0.8 & -- & 0.42 & 5.0 & 0.63 & 0.83 & 128.9 \\
& CCFM (Ours)              & \textbf{46.4} & \textbf{2.8} & \textbf{84.3} & \textbf{1.00} & 5.4 & 0.78 & 0.72 & \textbf{123.0} \\
\cmidrule(l){1-10}
\multirow{2}{*}{nuPlan\cite{nuplan}}
& SAFE-SIM \cite{safe-sim} & 42.1 & 2.4 & -- & 0.48 & 3.6 & 0.33 & 0.74 & \textbf{271.4} \\
& CCFM (Ours)              & \textbf{83.1} & \textbf{5.2} & \textbf{85.6} & \textbf{1.00}  & \textbf{1.7} & \textbf{0.33} & \textbf{1.00} & 356.3 \\
\midrule
\multicolumn{10}{l}{\textbf{Horizon = 200}} \\
\midrule
\multirow{3}{*}{nuScenes\cite{nuscenes}}
& CCDiff \cite{ccdiff}     & 11.3 & 0.7 & -- & 0.18 & \textbf{9.1} & 0.56 & \textbf{0.90} & 533.6 \\
& SAFE-SIM \cite{safe-sim} & 26.1 & 1.5 & -- & 0.41 & 11.4 & \textbf{0.45} & 0.90 & 322.5 \\
& CCFM (Ours)              & \textbf{59.9} & \textbf{4.0} & \textbf{85.4} & \textbf{1.00}  & 11.4 & 0.71 & 0.72 & \textbf{316.3} \\
\cmidrule(l){1-10}
\multirow{2}{*}{nuPlan\cite{nuplan}}
& SAFE-SIM \cite{safe-sim} & 50.6 & 3.0 & -- & 0.51 & 8.7 & \textbf{0.27} & 0.61 & \textbf{728.4} \\
& CCFM (Ours)              & \textbf{95.2} & \textbf{6.1} & \textbf{84.7} & \textbf{1.00} & \textbf{1.9} & 0.30 & \textbf{0.95} & 970.5 \\
\bottomrule
\end{tabular}
\label{tab:closed_loop_results}
\end{table}

\subsection{Comparison with SOTA Models}

First, we compare CCFM with STRIVE, CCDiff, and SAFE-SIM on nuScenes and nuPlan under different rollout horizons using the same Lane-graph ego planner \cite{strive}. 
As shown in Table \ref{tab:closed_loop_results}, under the practical setting of an 80-frame rollout horizon, CCFM achieves the highest collision rate on both datasets, reaching 46.4\% on nuScenes and 83.1\% on nuPlan. CCFM also achieves the highest severity of collision with the highest relative speed, demonstrating stronger safety-critical scenario generation capability. Extending the rollout horizon to 200 frames further improves the collision rate to 59.9\% on nuScenes, and 95.2\% on nuPlan, suggesting that a longer generation horizon evolves more safety-critical events with controllability. Despite some reduction in realism metrics, the substantial improvements in collision rate and severity make the trade-off acceptable.

\begin{table}[htbp]
\caption{Trade-off between criticality and realism. ADV.ADE and ADV.FDE denote the average displacement error and final displacement error of the adversarial agent, respectively. ADE, and FDE report the corresponding metrics over all generated trajectories. ADV.OR and OR report the off-road rate of the adversarial agent and reactive agents respectively. Metric definitions are detailed in the supplementary material.}
\centering
\setlength{\tabcolsep}{4pt}
\begin{tabular}{l | c|ccc|ccc}
\toprule
\textbf{Model}
& \makecell{\textbf{CR}\\ (\%) $\uparrow$} 
& \makecell{\textbf{ADV.OR}\\ (\%) $\downarrow$} 
& \makecell{\textbf{ADV.ADE}\\ (m) $\downarrow$}
& \makecell{\textbf{ADV.FDE}\\ (m) $\downarrow$}
& \makecell{\textbf{OR}\\ (\%) $\downarrow$} 
& \makecell{\textbf{ADE}\\ (m) $\downarrow$}
& \makecell{\textbf{FDE}\\ (m) $\downarrow$}
 \\

\midrule
STRIVE &20.8& 11.2 & 15.9 & 27.4 & 5.8 & 6.8 & 14.5\\
SAFE-SIM &25.8& \textbf{8.7} & \textbf{8.1} & \textbf{18.7} & \textbf{5.0} & \textbf{4.6} & \textbf{11.1} \\
CCFM &\textbf{46.4}& 10.0 & 12.0 & 24.9 & 5.4 & 5.9 & 13.3\\
\bottomrule
\end{tabular}
\label{Tab: trade-off}
\end{table}

To further explore the observed trade-off between realism and criticality, Table \ref{Tab: trade-off} details the realism evaluation of generated safety-critical scenarios at the scene level and the adversary agent level, respectively. The results show that CCFM achieves the highest collision rate, but maintains competitive behavior realism at both levels. Critically, the deviations are kinematically regulated, reflecting aggressive yet physically plausible maneuvers rather than unrealistic artifacts.

\subsection{Evaluation on Controllability and Diversity}
\subsubsection{Controllability.}
Quantitative results in Table \ref{tab:closed_loop_results} illustrate that only CCFM provides explicit collision-type controllability. CCFM achieves 84.3\% type match under the 80-frame rollout
 horizon and 85.4\% under the 200-frame rollout horizon, whereas STRIVE, CCDiff, and SAFE-SIM do not provide explicit type-level control. The consistent performance across the two rollout horizons indicates that CCFM can reliably steer generated scenarios toward the desired collision outcome, rather than merely increasing the overall collision rate.

\begin{figure}[b]
    
    \centering
    \includegraphics[width=1.00\linewidth]{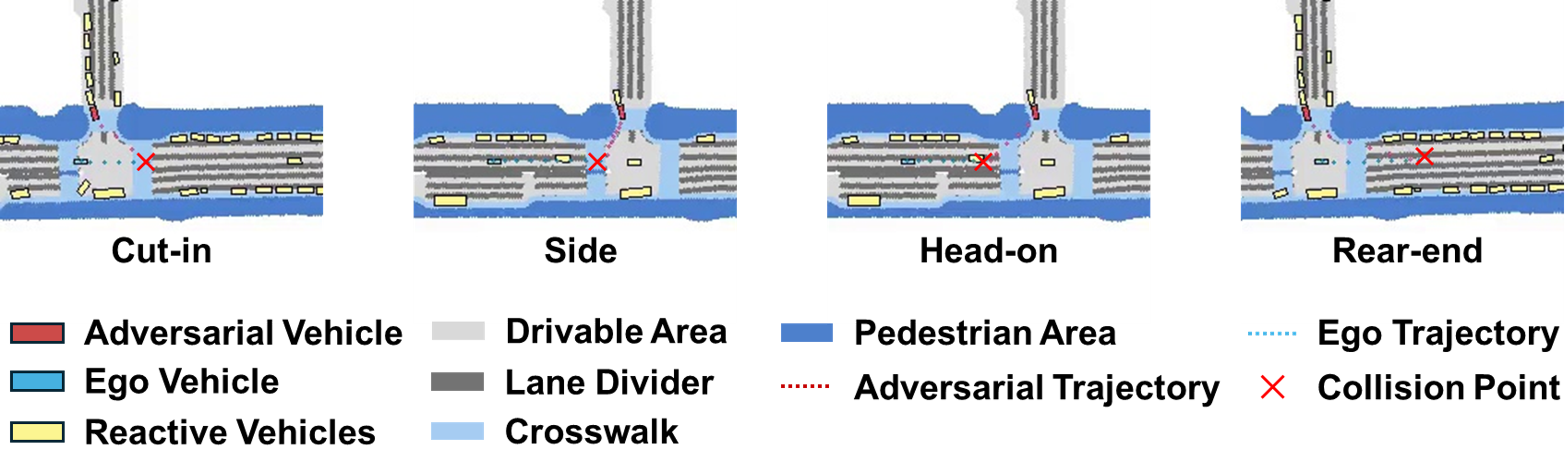}
    
    \caption{Qualitative visualization of collision-type controllability. Given an assigned ego--adversary pair and scene context, CCFM generates distinct scenarios by specifying target collision types.}
    \label{fig:visal}
    
\end{figure}

Furthermore, Fig. \ref{fig:visal} qualitatively illustrates the controllability of CCFM. Given identical initial scene contexts and a fixed ego-adversary pair, CCFM can generate distinct safety-critical interactions by altering the target constraint. In particular, the  \textit{Cut-in}, \textit{Head-on}, \textit{Side}, and \textit{Rear-end} examples are generated from the same scene context. These visualizations demonstrate that CCFM provides fine-grained, interpretable control over the final collision outcomes.

\subsubsection{Diversity.}
Compared to STRIVE, CCDiff, and SAFE-SIM, our method achieves the second-best contact-region entropy with EN=0.89, indicating a relatively diverse distribution over Rear, Front, and Side ego contact regions, as shown in Table \ref{tab:closed_loop_results diversity}. 
Although CCDiff obtains the highest entropy score of EN=0.97, its collision rate remains much lower at only 4.0\%.  In addition, CCFM achieves a larger standard deviation in relative speed (SD = 4.9 m/s) and relative angle (HD = $30.9^\circ$), indicating richer diversity in collision kinematics and interaction geometry. The highest DS = 0.96 suggests that CCFM does not collapse to a single collision configuration, but is able to generate safety-critical scenarios with various impact forms and motion characteristics.

\begin{table}[t]
\caption{Comparison of diversity results with baseline models. We evaluate the distribution of three contact regions at collision by EN. Meanwhile, the standard deviation SD and HD reflect the diversity of relative speed and heading angle. $^\dagger$ denotes relative composite scores normalized within this comparison table.}
\centering

\setlength{\tabcolsep}{6pt}
\begin{tabular}{l| c c c c | c c | >{\columncolor{gray!12}}c}
\toprule
\textbf{Model} 
& \makecell{\textbf{Rear}\\(\%)} 
& \makecell{\textbf{Front}\\(\%)} 
& \makecell{\textbf{Side}\\(\%)} 
& \makecell{\textbf{EN}\\$\uparrow$}
& \makecell{\textbf{SD}\\($m/s$) $\uparrow$} 
& \makecell{\textbf{HD}\\(${^\circ}$) $\uparrow$}
& \makecell{\textbf{DS$^\dagger$}\\$\uparrow$}  \\
\midrule

STRIVE \cite{strive}    
& 18.4 & 24.2 & 57.4 & 0.86 
& $1.9 \pm 2.8$ 
& $2.2 \pm 12.9$ 
& 0.62 \\

CCDiff \cite{ccdiff}  
& 26.6 & 28.2 & 45.2 & \textbf{0.97}
& $0.3 \pm 1.7$ 
& $1.5 \pm 14.1$ 
& 0.59 \\

SAFE-SIM \cite{safe-sim}  
& 10.2 & 28.6 & 61.2 & 0.81 
& $0.8 \pm 2.5$ 
& $2.4 \pm 15.8$ 
& 0.61 \\

CCFM (Ours)   
& 50.5 & 14.2 & 35.3 & 0.89 
& $\mathbf{2.8} \pm \mathbf{4.9}$ 
& $\mathbf{12.3} \pm \mathbf{30.9}$ 
& \textbf{0.96} \\

\bottomrule
\end{tabular}
\label{tab:closed_loop_results diversity}
\end{table}

\subsection{Assessment Across Ego Planners}

\begin{table}[b]
\caption{Closed-loop simulation results using different ego planners for assessing CCFM planner-agnostic controllability and effectiveness.}
\centering
\setlength{\tabcolsep}{6pt}

\begin{tabular}{l| c c| c |c c}
\toprule
\textbf{Planner} 
& \textbf{CR} (\%) $\uparrow$ 
& \textbf{MS} ($m/s$) $\uparrow$ 
& \textbf{TM}(\%) $\uparrow$ 
& \textbf{OR}(\%) $\downarrow$  
& \textbf{RM}$\downarrow$ \\
\midrule
IDM \cite{IDM}              & \textbf{51.6} & \textbf{3.3} & 82.7          & \textbf{5.1} & 0.81 \\
PDM \cite{pcm-closed}       & 36.1          & 2.6          & 81.2          & 8.5          & 0.85 \\
Lane-graph \cite{strive}    & 46.4          & 2.8          & \textbf{84.3} & 5.4          & \textbf{0.78} \\
\bottomrule
\end{tabular}

\label{tab:closed_loop_results with different planner}
\end{table}

We evaluate the proposed CCFM framework with three ego planners: IDM \cite{IDM}, PDM \cite{pcm-closed}, and the lane-graph method used in STRIVE \cite{strive}. As shown in Table \ref{tab:closed_loop_results with different planner}, CCFM consistently achieves high collision rate, and strong type match performance across different ego planners, indicating planner-agnostic controllability and robust effectiveness. Although CCFM achieves a relatively lower collision rate with the PDM planner, this is expected because PDM's structured motion planning and explicit collision-avoidance mechanism allow the ego vehicle to execute more robust evasive maneuvers. 

Meanwhile, the variation across ego planners further suggests that stronger planning algorithms can avoid a subset of the generated safety-critical interactions. However, our current evaluation does not explicitly distinguish avoidable collisions from physically unavoidable ones. Future work will incorporate avoidability analysis to better quantify the planner relevance of generated scenarios, and explore their use as targeted training data to enhance the robustness and generalization of downstream planning models.

\subsection{Ablation Study}
We conduct a series of ablation studies to evaluate the effectiveness of three key components: Heuristic Collision Selector, Collision-Type-specific Constraints, and Collision-Constrained Gauss–Newton Projection of our CCFM framework in safety-critical closed-loop simulation.

\subsubsection{Effectiveness of Heuristic Collision Selector.}
Table \ref{tab:ablation HCS} shows that HCS achieves the best overall performance compared to fixed single-type selectors. 
Although individual fixed strategies may perform well on a single metric, such as higher EN for \textit{Rear-end only} or higher MS for \textit{Cut-in only}, they suffer from limited controllability or realism. 
In contrast, HCS provides a more effective trade-off among criticality, diversity, controllability, and realism. These results demonstrate that adaptively selecting the adversarial agent and collision type based on reachability, geometric alignment, and lane-graph legality provides a feasible basis for satisfying the subsequent collision-type-specific constraints.

\begin{table}[t]
\caption{Ablation study on the Heuristic Collision Selector (HCS). We compare fixed single-type selectors with the HCS, which adaptively assigns ego-adversary pair and feasible collision type. $^\dagger$ denotes relative composite scores normalized within this comparison table.}
\centering
\setlength{\tabcolsep}{3pt}

\begin{tabular}{l| c c c >{\columncolor{gray!12}}c | c >{\columncolor{gray!12}}c | c c >{\columncolor{gray!12}}c | >{\columncolor{gray!12}}c}
\toprule
\centering
\setlength{\tabcolsep}{5pt}

\textbf{HCS} 
& \makecell{\textbf{CR}\\(\%) $\uparrow$} 
& \makecell{\textbf{TM}\\(\%) $\uparrow$}
& \makecell{\textbf{MS}\\($m/s$) $\uparrow$}
& \makecell{\textbf{CS}$^\dagger$\\$\uparrow$}
& \makecell{\textbf{EN}\\$\uparrow$}
& \makecell{\textbf{DS$^\dagger$}\\$\uparrow$}
& \makecell{\textbf{OR}\\(\%) $\downarrow$} 
& \makecell{\textbf{RM}\\$\downarrow$}
& \makecell{\textbf{RS$^\dagger$}\\$\uparrow$} 
& \makecell{\textbf{CWS$^\dagger$}\\$\uparrow$}\\
\midrule
Rear-end only   
& 30.2         
&70.7 
& 2.2
& 0.51
& \textbf{0.96}
& 0.69
& 7.0          
& 0.86
& 0.84
&0.68\\

Head-on only    
& 27.7          
& 16.4           
& 2.7
& 0.36
& 0.72
& 0.83
& 7.5          
& 0.85
&0.82
&0.67\\

Side only       
& 31.4          
&14.5
& 3.1
& 0.41
& 0.64
& 0.75
& 6.7          
& 0.87
& 0.85
&0.67\\

Cut-in only     
& 40.0 
& 4.6           
& \textbf{4.4}
& 0.52
& 0.79
& 0.82
&  5.5       
& 0.80
& 0.98
&0.77\\

HCS             
& \textbf{46.4}          
& \textbf{84.3}          
& 2.8
& \textbf{0.82}
& 0.89
& \textbf{0.84}
& \textbf{5.4}  
& \textbf{0.78}
& \textbf{1.00}
&\textbf{0.89}\\
\bottomrule
\end{tabular}
\label{tab:ablation HCS}
\end{table}

\subsubsection{Effectiveness of Collision Type-specific Constraint Components.}
As shown in Table \ref{tab:ablation constraint set}, each constraint component contributes differently to controllability, diversity, realism, and overall generation quality. Using only the contact point constraint already produces safety-critical scenarios with a collision rate of 51.6\% and a type match of 82.1\%, showing that $h_{\mathrm{cnt},t}$ is effective for triggering collisions at the desired contact region. Adding the relative heading constraint slightly reduces the collision rate to 48.4\%, but improves the type match to 83.8\% and diversity score to 0.92, indicating that $h_{\mathrm{hdg},t}$ provides stronger geometric controllability over the target collision type and greater diversity respectively. When the full set of constraints is applied, the type match further increases to 84.3\%, and the relative speed increases to 2.8 m/s, suggesting that the severity constraint $h_{\mathrm{svt},t}$ encourages more intense safety-critical interactions. Meanwhile, the full model achieves the highest criticality score, entropy score, diversity score, RS, and composite weighted score, demonstrating the best overall balance among criticality, diversity, realism, and controllability.

\begin{table}[t]
\caption{Ablation study on different components in collision-type-specific constraint. We evaluate the contribution of contact point constraint $h_\mathrm{cnt}$, relative heading constraint $h_\mathrm{hdg}$, and severity constraint $h_\mathrm{svt}$ to safety-critical scenario generation in closed-loop simulation. $^\dagger$ denotes relative composite scores normalized within this comparison table.}
\centering

\setlength{\tabcolsep}{4pt}
\begin{tabular}{c c c | c c c >{\columncolor{gray!12}}c | c >{\columncolor{gray!12}}c | >{\columncolor{gray!12}}c |>{\columncolor{gray!12}}c}
\toprule
$h_{\mathrm{cnt},t}$ 
& $h_{\mathrm{hdg},t}$ 
& $h_{\mathrm{svt},t}$  
& \makecell{\textbf{CR}\\(\%) $\uparrow$}
& \makecell{\textbf{TM}\\(\%) $\uparrow$}
& \makecell{\textbf{MS}\\($m/s$) $\uparrow$}
& \makecell{\textbf{CS}$^\dagger$\\$\uparrow$}
& \makecell{\textbf{EN}\\$\uparrow$}
& \makecell{\textbf{DS$^\dagger$}\\$\uparrow$}
& \makecell{\textbf{RS}$^\dagger$\\$\uparrow$}
& \makecell{\textbf{CWS}$^\dagger$\\$\uparrow$} \\
\midrule
$\checkmark$    
& 
&                 
& \textbf{51.6}          
& 82.1          
& 2.3         
& 0.91
& 0.85
& 0.90
& 0.94
& 0.92\\

$\checkmark$    
& $\checkmark$
&     
& 48.4          
& 83.8
& 2.6        
& 0.94
& 0.86
& 0.92
& 0.94
& 0.93\\

$\checkmark$    
& $\checkmark$
& $\checkmark$
& 46.4 
& \textbf{84.3}          
& \textbf{2.8}  
& \textbf{0.96}
&\textbf{0.89}
& \textbf{0.96}
& \textbf{0.96}
& \textbf{0.96}\\
\bottomrule
\end{tabular}
\label{tab:ablation constraint set}
\end{table}

\begin{table}[b]
\centering
\caption{Ablation study comparing soft collision guidance and manifold projection. Starting from the baseline Flow Matching (FM) model, we evaluate the effect of adding soft collision guidance $J_\mathrm{col}$ and the proposed Gauss-Newton (GN) manifold projection. $^\dagger$ denotes relative composite scores normalized within this comparison table.}
\setlength{\tabcolsep}{4pt}
\begin{tabular}{l| c c >{\columncolor{gray!12}}c | c >{\columncolor{gray!12}}c | c c >{\columncolor{gray!12}}c | >{\columncolor{gray!12}}c}
\toprule
\textbf{Model} 
& \makecell{\textbf{CR}\\(\%) $\uparrow$}
& \makecell{\textbf{MS}\\($m/s$) $\uparrow$}
& \makecell{\textbf{CS}$^\dagger$\\$\uparrow$}
& \makecell{\textbf{EN}\\$\uparrow$}
& \makecell{\textbf{DS}$^\dagger$\\$\uparrow$}
& \makecell{\textbf{OR}\\(\%) $\downarrow$}
& \makecell{\textbf{RM}\\$\downarrow$}
& \makecell{\textbf{RS}$^\dagger$\\$\uparrow$}
& \makecell{\textbf{CWS}$^\dagger$\\$\uparrow$} \\
\midrule

FM     
& 8.2 
& 0.7  
& 0.21
& 0.63 
& 0.48
& \textbf{2.3} 
& \textbf{0.77}  
&  \textbf{1.00}
&  0.56\\

FM $+ J_\mathrm{col}$   
& 10.2 
& 0.8
& 0.25
& 0.70 
& 0.43 
& 6.2 
& 0.92  
&  0.60
&  0.43\\

FM $+ \text{GN}$  
& \textbf{46.4} 
& \textbf{2.8} 
& \textbf{1.00}
& \textbf{0.89}
& \textbf{0.97}
&  5.4
& 0.78 
&  0.71
&  \textbf{0.89}\\

\bottomrule
\end{tabular}
\label{tab:ablation GN projection}
\end{table}

\subsubsection{Effectiveness of Manifold Projection versus Soft Guidance.}
As shown in Table \ref{tab:ablation GN projection}, the soft collision guidance $J_{\mathrm{col}}$, implemented following SAFE-SIM \cite{safe-sim}, only marginally improves the collision rate from 8.2\% to 10.2\% compared to the baseline Flow Matching model. 
In contrast, the proposed GN manifold projection substantially increases the collision rate to 46.4\% with fewer denoising steps. Beyond collision rate, GN projection also leads to more critical and diverse collision outcomes, achieving the highest relative speed of 2.8 m/s, and the highest diversity score of 0.97. Although its realism score is lower than the FM baseline, the overall CWS reaches 0.89, significantly outperforming both FM and soft guidance.  Overall, the results suggest that hard projection onto the collision-constrained feasible set is significantly more effective than soft guidance for structured safety-critical generation. However, the final outcome still depends on the road topology, traffic context, projection feasibility, and the ego planner's closed-loop response. Therefore, CCFM explicitly targets collision configurations, but a collision is realized only when the constraint can be satisfied within the evolving interactive scenario.

\section{Conclusion}

In this paper, we developed a Collision-Constrained Flow Matching (CCFM) method for controllable safety-critical scenario generation. 
Extensive experiments on nuScenes closed-loop simulation demonstrated that CCFM generates more safety-critical scenarios than prior methods, while achieving interpretable collision-type controllability, diverse collision outcomes, and competitive realism. In particular, our results showed that hard projection is considerably more effective than soft guidance for enforcing structured collision constraints in generation. Future work will focus on using CCFM-generated safety-critical scenarios to stress-test and improve end-to-end AV systems, as well as integrating it into sim-to-real safety-critical scenario generation for improving real-world validation.

\section*{Acknowledgments}

Qin receives funding from the Rural Safe Efficient Advanced Transportation (R-SEAT) Center, a Tier-1 University Transportation Center (UTC) funded by the United States Department of Transportation (USDOT), through agreement number 69A3552348321. The contents of this paper reflect the views of the authors. USDOT assumes no liability for the contents or use thereof.

\bibliographystyle{splncs04}
\bibliography{main}

\clearpage
\appendix

\title{Supplementary Material: CCFM: Collision-Constrained Flow Matching for Safety-Critical Scenario Generation} 

\titlerunning{CCFM}

\author{Ke Li\inst{1}\orcidlink{0009-0001-4958-3302} \and
Kaidi Liang\inst{1}\orcidlink{0009-0001-9129-2744} \and
Yuxin Ding\inst{2}\orcidlink{0009-0005-3629-1908} \and Debojyoti Biswas\inst{2} \orcidlink{0000-0002-8842-0207} \and Xianbiao Hu\inst{2}  \orcidlink{0000-0002-0149-1847} \and Ruwen Qin \inst{1}$^*$  \orcidlink{0000-0003-2656-8705}}

\authorrunning{K.~Li et al.}

\institute{Stony Brook University, Stony Brook, NY 11794, USA \and
Pennsylvania State University, University Park, PA 16802, USA\\
\email{\{ke.li.1, kaidi.liang, ruwen.qin\}@stonybrook.edu}\\
\email{\{ymd5170, dbb5917, xbhu\}@psu.edu}\\
$^*$ Corresponding author\\
\url{https://github.com/KELISBU/CCFM}}

\maketitle

\appendix

The supplementary material provides details on modeling the heuristic collision selector, deriving collision-type-specific constraints, the Gauss-Newton projection algorithm, evaluation metrics, and additional experimental results that are omitted from the main paper due to page limits.

\section{Heuristic Collision Selector Modeling}
\label{sec:HCS_SM}

\newcommand{\Srch}{S_{\mathrm{rch},t}}
\newcommand{\Sgeo}{S_{\mathrm{geo},t}}
\newcommand{\Slgt}{S_{\mathrm{lgt},t}}
\newcommand{\psirch}{\psi_{\mathrm{rch}}}
\newcommand{\psigeo}{\psi_{\mathrm{geo}}}
\newcommand{\psilgt}{\psi_{\mathrm{lgt}}}
\newcommand{\ego}{\mathrm{ego}}
\newcommand{\pego}{\mathbf{p}_{\ego,t}}
\newcommand{\pit}{\mathbf{p}_{i,t}}

\subsection{Notation}
\label{subsec:Notation}

The major symbols and notations used in modeling the Heuristic Collision Selector (HCS) are summarized in Table \ref{tab:notation}.

\begin{table}[t]
\centering
\caption{Symbols for Modeling HCS}
\renewcommand{\arraystretch}{1.3}
\begin{tabular}{lp{4in}}
\toprule
\textbf{Symbol} & \textbf{Description} \\
\midrule
$\vartheta_{i,t},\;\vartheta_{\mathrm{ego},t}$ & Heading angles of agent $i$ and the ego vehicle \\
$b_{i,t}$
  & Relative bearing of agent $i$ with respect to the ego vehicle at $t$ \\
$d_{i,t}$
  & Heading alignment between agent $i$ and the ego vehicle at $t$\\
$g_t$
  & Proximity-aware gate for activating the geometry constraint \\
$v_{i,t},\;v_{\mathrm{ego},t}$
  & Speeds of agent $i$ and the ego vehicle \\
$L_{\mathrm{ego}},W_{\mathrm{ego}}$
  & The length and width of ego vehicle \\
$S_{i,t}$&
Collision score for agent $i$ at $t$\\
$S_{\mathrm{rch},t}(i)$& Reachability score for agent $i$ at $t$\\
$S_{\mathrm{geo},t}(i,m)$& Geometry score of agent $i$ in type $m$ collision at $t$ \\
$S_{\mathrm{lgt},t}(i,m)$& Legality score of agent $i$ in type $m$ collision at $t$\\
$\mathbf{a}_{i,t:t+T-1}$& Action sequence\\
$\mathbf{c}_{i,t},\;\mathbf{c}_{\mathrm{ego},t}$
  & Lane-centerline contexts of agent $i$ and the ego vehicle \\
$\mathbf{p}_{i,t},\;\mathbf{p}_{\mathrm{ego},t}$
  & World-frame positions of agent $i$ and the ego vehicle at time $t$ \\
$\mathbf{pos}_{i,t},\mathbf{pos}_m$& Pose feature vector of agent $i$ at $t$, and the target pose in type $m$ collision\\
$\mathbf{n}_{i,t}$, $\mathbf{n}_{\mathrm{ego},t}$
  & Unit heading vectors of agent $i$ and the ego vehicle \\
$\mathbf{n}^\mathrm{rel}_{i,t}$
  & Unit relative position vector from agent $i$ to the ego vehicle \\
$\mathbf{n}_{\mathrm{ego},t}^{\perp}$
  &Lateral axis of the ego vehicle \\
$\mathbf{tpl}_{i,t}, \mathbf{tpl}_\mathrm{m}$& Topological relationship between  agent $i$ and the ego vehicle at $t$ and the typical relationships\\ 
$\mathbf{v}_{i,t},\;\mathbf{v}_{\mathrm{ego},t}$
  & Velocities of agent $i$ and the ego vehicle at time $t$ \\
$\mathcal{M}$
  & $=\{\text{Rear-end, Side, Cut-in, Head-on}\}$, Set of collision types\\
$\mathcal{L}$
  & $=\{$intersection, merging, same lane, nearby lanes, others$\}$, Set of topological relationships \\
\bottomrule
\end{tabular}
\label{tab:notation}
\end{table}

\subsection{Detailed Score Functions for HCS}

HCS is based on the estimation of three similarity-based scores, which are delineated as follows.

\subsubsection{Reachability Score.}
This score for agent $i$ at timestep $t$ is estimated using a similarity function $\psirch(\cdot)$, an isotropic Gaussian kernel centered on the ego vehicle:

\begin{equation}
    \Srch(i)
    = \psirch\!\left(\pit,\;\pego\right)
    = \exp\!\left(-\frac{\left\|\pit - \pego\right\|^{2}}{\sigma^{2}}\right),
    \label{eq:s_reach}
\end{equation}
where $\mathbf{p}_{i,t}$ is the position of the candidate agent $i$ at timestep $t$;  $\mathbf{p}_{\mathrm{ego},t}$ is the position of the ego vehicle at $t$; and the bandwidth $\sigma=50\,\text{m}$ is the shape parameter of the function. Eq. (\ref{eq:s_reach}) means that the reachability score decays exponentially when the Euclidean distance between agent $i$ and the ego vehicle increases, approaching zero beyond the typical interaction range.

\begin{figure}[t]
    
    \centering
    \includegraphics[width=1.0\linewidth]{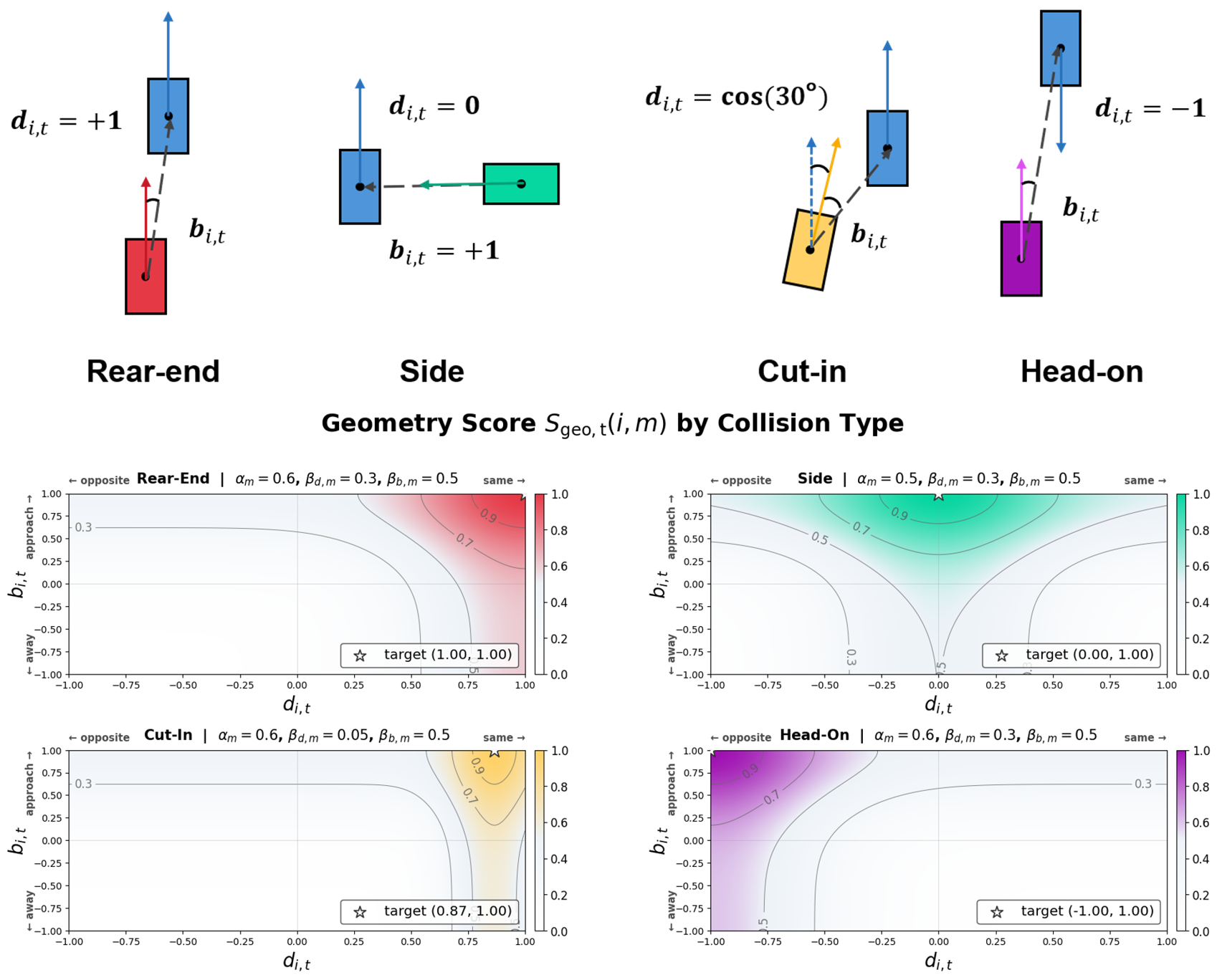}
    
    \caption{Illustration of the geometry score $S_{\mathrm{geo},t}(i,m)$ for different collision types. For each collision type, the top row shows the target geometric configuration in terms of heading alignment $d_{i,t}$ and bearing feature $b_{i,t}$, while the bottom visualizes the corresponding geometry score landscape over the $(d_{i,t},b_{i,t})$ space. The star marks the target feature vector $\mathbf{pos}_{m}$ for each collision type $m$.}
    \label{fig:HCS:geo}
    
\end{figure}

\subsubsection{Geometry Score.}
As illustrated in Fig.~\ref{fig:HCS:geo}, computing this score for candidate agent $i$ under collision type $m$ at timestep $t$ requires first determining the heading alignment between the two vehicles, $d_{i,t}$, and the relative bearing of the agent $i$ with respect to the ego vehicle, $b_{i,t}$: 
\begin{equation}
d_{i,t} = \mathbf{n}_{i,t} \cdot \mathbf{n}_{\mathrm{ego},t},
\quad
b_{i,t} = \mathbf{n}^{\mathrm{rel}}_{i,t} \cdot \mathbf{n}_{i,t},
\end{equation}
where $\mathbf{n}_{i,t} = [\cos\vartheta_{i,t}, \sin\vartheta_{i,t}]$ and $\mathbf{n}_{\mathrm{ego},t} = [\cos\vartheta_{\mathrm{ego},t}, \sin\vartheta_{\mathrm{ego},t}]$ denote the unit heading vectors of agent $i$ and the ego vehicle, respectively, and the unit vector
\begin{equation}
\mathbf{n}^{\mathrm{rel}}_{i,t} = \frac{\mathbf{p}_{\mathrm{ego},t} - \mathbf{p}_{i,t}}{\left\| \mathbf{p}_{\mathrm{ego},t} - \mathbf{p}_{i,t} \right\|}.
\end{equation}
represents the relative position direction from agent $i$ to the ego vehicle.

The dot product $d_{i,t}\in[-1,1]$ encodes the heading alignment: 1 represents the same direction, -1 denotes the opposite direction, and 0 indicates the perpendicular direction. The dot product $b_{i,t}\in[-1,1]$ encodes the
bearing of agent $i$ relative to the ego vehicle: 1 means approaching directly, -1 represents directly moving away, and 0 denotes a relative lateral position. Together, they form the relative pose feature vector:
\begin{equation}
    \mathbf{pos}_{i,t}\!\left(\mathbf{p}_{i,t}, \vartheta_{i,t}, \mathbf{p}_{\mathrm{ego},t}, \vartheta_{\ego,t}\right)
    = \bigl[d_{i,t},\;b_{i,t}\bigr].
\end{equation}

Each collision type $m$ is associated with a specific target pose
$\mathbf{pos}_m = [d_{m},b_{m}]$, identified as the star symbol illustrated in Fig. \ref{fig:HCS:geo}. The similarity function $\psigeo(\cdot)$ is a
weighted sum of Gaussian kernels over the two pose features:
\begin{equation}
    \Sgeo(i,m) = \psigeo\!\left(\mathbf{pos}_{i,t},\;\mathbf{pos}_m\right)\\
    = \left[
            \begin{aligned}
                &\alpha_m\\
                &1-\alpha_m
            \end{aligned}
        \right]^\mathrm{T}
    \left[
    \begin{aligned}
        \exp(-(d_{i,t}-d_m)^2/\beta_{d,m})\\
        \exp(-(b_{i,t}-b_m)^2/\beta_{b,m})\\
    \end{aligned}
    \right]
    \label{eq:s_geo_general}
\end{equation}
where $\alpha_m\in[0,1]$ controls the relative importance between the two pose features; and the bandwidth parameters $\beta_{d,m}$ and $\beta_{b,m}$ control the tolerance of the similarity function with respect to deviations from their target values. Each Gaussian kernel in eq. (\ref{eq:s_geo_general}) produces a high response when the corresponding feature is close to its target value and decays smoothly as the deviation increases. The final geometric similarity score $S_{\mathrm{geo},t}(i,m)$ is obtained as a weighted combination of the two responses, capturing both heading alignment and relative bearing of agent $i$ with respect to the ego vehicle in a collision of type $m$.

The collision-type-specific parameters for calculating the geometry score are summarized in Table \ref{tab:geo_params} and also displayed in Fig. \ref{fig:HCS:geo}.

\begin{table}[h]
\setlength{\tabcolsep}{10pt}
\centering
\caption{Geometry score parameters by collision type}
\label{tab:geo_params}
\begin{tabular}{l|rrrrr}
\toprule
$m$ & $d_{m}$ & $\beta_{d,m}$ & $b_{m}$ & $\beta_{b,m}$ & $\alpha_m$ \\
\midrule
Rear-end & $+1$  & 0.3     & $+1$                   & $0.5$ & $0.6$ \\
Side     & $0$           & $0.3$ & $+1$                    & $0.5$ & $0.5$ \\
Cut-in   & $\cos 30^{\circ}$ & $0.05$ & $+1$ & $0.5$ & $0.6$ \\
Head-on  & $-1$          & $0.3$ & $+1$                   & $0.5$ & $0.6$ \\
\bottomrule
\end{tabular}
\end{table}

\subsubsection{Legality Score.}
The score for agent $i$ at timestep $t$ measures the likelihood that the agent can cause a collision of type $m$, according to the observed topological relationship between agent $i$ and the ego vehicle, $\mathbf{tpl}_{i,t}$, in comparison to the typical topological configuration associated with the collision type $m$, $\mathbf{tpl}_m$.

Following SAFE-SIM \cite{safe-sim}, the topological relationship between agent $i$ and the ego vehicle at timestep $t$ is derived from their contexts, positions, and headings, and represented as a one-hot vector:
\begin{equation}
\mathbf{tpl}_{i,t} \left(
    \mathbf{c}_{i,t}, \mathbf{p}_{i,t}, \vartheta_{i,t},
    \mathbf{c}_{\mathrm{ego},t}, \mathbf{p}_{\mathrm{ego},t}, \vartheta_{\mathrm{ego},t}
\right)\in\{0,1\}^{|\mathcal{L}|}.
\end{equation}
where:
\begin{equation}
    \mathcal{L}=\{\texttt{same\_lane}, \texttt{nearby\_lanes}, \texttt{merging}, \texttt{intersection},\texttt{others}\}.
\end{equation}

To model the typical topological configurations associated with
different collision types, we define a compatibility matrix in Table \ref{tab:phi_compat}, $\mathbf{M}_{\mathrm{lgt}} \in \mathbb{R}^{|\mathcal{L}|\times |\mathcal{M}|}$, 
whose entries take values in the range $[0,1]$ and specify the typicality of each relationship for each collision type. This matrix can be calibrated using historical data in crash reporting systems.

Accordingly, the typical topological configuration for crash type $m$ is given by
\begin{equation}
    \mathbf{tpl}_m = \mathbf{M}_{\mathrm{lgt}} \mathbf{e}_m,
\end{equation}
where $\mathbf{e}_m$ is the one-hot indicator vector for collision type $m$.

The legality score is then defined as:
\begin{equation}
    S_{\mathrm{lgt},t}(i, m) 
    = \psi_{\mathrm{lgt}}\!\left(\mathbf{tpl}_{i,t},\; \mathbf{tpl}_m\right) = \mathbf{tpl}_{i,t}^\mathrm{T}\, \mathbf{tpl}_m,
    \label{eq:s_lgt}
\end{equation}
where the similarity function $\psi_\mathrm{lgt}$ is a dot product operator that selects the compatibility score corresponding to the observed topological relationship. 

\begin{table}[htb]
\centering
\setlength{\tabcolsep}{10pt}
\caption{Base compatibility scores for each topological relationship and collision type}
\label{tab:phi_compat}

\begin{tabular}{c|l|cccc}
\toprule
 \multicolumn{2}{l}{}  & \multicolumn{4}{c}{Collision Type} \\
\cline{3-6}
 \multicolumn{2}{l}{} 
 & Side & Head-on & Cut-in & Rear-end \\
\midrule

\multirow{5}{*}{\makecell{Topological\\Relationship}}
 & \texttt{intersection}  & 0.95 & 0.70 & 0.30 & 0.30 \\
 & \texttt{merging}       & 0.60 & 0.10 & 0.95 & 0.40 \\
 & \texttt{same\_lane}    & 0.05 & 0.05 & 0.05 & 0.95 \\
 & \texttt{nearby\_lanes} & 0.40 & 0.20 & 0.70 & 0.60 \\
 & \texttt{others}                 & 0.01 & 0.01 & 0.01 & 0.01 \\

\bottomrule
\end{tabular}
\end{table}


\subsubsection{The Weighted Score.}
The weights used to combine the three scores are $w_{\mathrm{rch}}$, $w_{\mathrm{geo}}$, and $w_{\mathrm{lgt}}$, which sum to one. They are set to be equal, making $S_t(i,m)$ the simple average of the three scores. The score function is evaluated over all candidate agents and collision types to identify the optimal pair of adversarial vehicle and collision type that maximizes the score:

\begin{equation}
(i^{*},m^{*})
= \operatorname*{\argmax}_{i\in\{1,\dots,N-1\},\;m\in\mathcal{M}} S_t(i,m).
\end{equation}

Based on this, the trajectory of the adversarial agent $i^\ast$ is generated using the proposed Collision-Constrained Flow Matching (CCFM) method to induce a collision with the ego vehicle of the target type $m^\ast$.

\section{Collision-Type-Specific Constraints}
\label{sec:App Collision-type-specific Constraints}

We define a set of constraints for each representative collision type and imposes them on the predicted residual $\mathbf{h}_t$ to guide trajectory generation for the adversarial agent during closed-loop simulation. Specifically, $\mathbf{h}_t$ is measured at $T_\mathrm{col}$, the target future timestep at which a collision of the target type $m^\ast$ will occur. Its estimation is delineated in Sec. \ref{sec:dynamic_tcol}. 

All collision-type-specific constraints are imposed at the target collision timestep,  $t+T_{\mathrm{col}}$, and are derived from the predicted states of the adversarial vehicle $i^\ast$ and the ego vehicle for a collision of type $m^\ast$. For notational simplicity, we define
\begin{equation}
\begin{aligned}
&\hat{\mathbf{p}}_{i^\ast} \triangleq \mathbf{p}_{i^\ast,t+T_{\mathrm{col}}},\quad
\hat{\mathbf{n}}_{i^\ast} \triangleq \mathbf{n}_{i^\ast,t+T_{\mathrm{col}}},\quad
\hat{v}_{i^\ast} \triangleq v_{i^\ast,t+T_{\mathrm{col}}},\quad \hat{\vartheta}_{i^\ast}\triangleq \vartheta_{i^\ast,t+T_\mathrm{col}}\\
&\hat{\mathbf{p}}_{\mathrm{ego}} \triangleq \mathbf{p}_{\mathrm{ego},t+T_{\mathrm{col}}},\quad
\hat{\mathbf{n}}_{\mathrm{ego}} \triangleq \mathbf{n}_{\mathrm{ego},t+T_{\mathrm{col}}},\quad
\hat{v}_{\mathrm{ego}} \triangleq v_{\mathrm{ego},t+T_{\mathrm{col}}},\quad\hat{\vartheta}_{\mathrm{ego}}\triangleq \vartheta_{\mathrm{ego},t+T_\mathrm{col}}
\end{aligned}
\label{eq:states_Tcol}
\end{equation}
following the notation of Section \ref{subsec:Notation}. 

We define dissimilarity measures based on the predicted states in Eq. (\ref{eq:states_Tcol}) and impose constraints on them when generating the adversarial agent's action sequence. These measurements are introduced below. 

\subsection{Contact Constraint} 
At timestep $t$, we define $l_{\mathrm{cnt},t}$ as the predicted distance between the target contact points of the adversarial agent $i^\ast$ and the ego vehicle for a collision of type $m^\ast$ after $T_{\mathrm{col}}$ steps, calculated as:
\begin{equation}
l_{\mathrm{cnt},t} =
    \left\{
    \begin{aligned}
        &\min(\|\hat{\mathbf{p}}_{i^\ast}^\text{r}-\hat{\mathbf{p}}_{\mathrm{ego}}^\text{f}\|,\|\hat{\mathbf{p}}_{i^\ast}^\text{f}-\hat{\mathbf{p}}_{\mathrm{ego}}^\text{r}\|),\; \text{if } m^\ast = \text{Rear-end};\\
        & \|\hat{\mathbf{p}}_{i^\ast}^\text{f}-\hat{\mathbf{p}}_{\mathrm{ego}}^\text{f}\|,\; \text{if } m^\ast = \text{Head-on};\\
        &\|\hat{\mathbf{p}}_{i^\ast}^\text{c}-\hat{\mathbf{p}}_{\mathrm{ego}}^\text{s}\|,\; \text{if }m^\ast= \text{Side};\\
        &\|\hat{\mathbf{p}}_{i^\ast}^\text{c}-\hat{\mathbf{p}}_{\mathrm{ego}}^\text{f}\|,\; \text{if }m^\ast= \text{Cut-in};
    \end{aligned}\\
    \right.
\end{equation}
where the superscripts $\mathrm{f}$, $\mathrm{r}$, $\mathrm{s}$, and $\mathrm{c}$ denote
the front, rear, side, and center points of the vehicles, respectively. 

We define $\tilde{l}_\mathrm{cnt}$ as the threshold distance between the contact points for a collision of type $m^\ast$, as summarized in Table \ref{tab:constraints}. Accordingly, a dissimilarity function calculates the contact residual:
\begin{equation}
    h_{\mathrm{cnt},t} = \phi_{\mathrm{cnt}}(l_{\mathrm{cnt},t},\,\tilde{l}_{\mathrm{cnt}})
    = w_{\mathrm{cnt}} \cdot \mathrm{ReLU}(l_{\mathrm{cnt},t} - \tilde{l}_{\mathrm{cnt}}),
    \label{eq:phi_cnt}
\end{equation}
where $\mathrm{ReLU}$ denotes the Rectified Linear Unit function and $w_\mathrm{cnt}$ is a scaling factor. $h_{\mathrm{cnt},t}$ measures how much the predicted contact distance exceeds the threshold defined for that collision type.

\subsection{Heading Constraint}

At timestep $t$, we also define $l_{\mathrm{hdg},t}$ as the predicted heading alignment between the adversarial agent $i^\ast$ and the ego vehicle in a collision of type $m^\ast$ after $T_\mathrm{col}$ steps, calculated as:
\begin{equation}
    l_{\mathrm{hdg},t}=\hat{d}_{i^\ast} =
\hat{\mathbf{n}}_{i^*}\cdot\hat{\mathbf{n}}_\mathrm{ego}.
\end{equation}

$\tilde{l}_{\mathrm{hdg}} = [\underline{d},\, \overline{d}]$
is the target alignment interval in a collision of type $m^\ast$, specified in Table \ref{tab:constraints}. Then, a dissimilarity function calculates the heading residual:
\begin{equation}
\begin{aligned}
    h_{\mathrm{hdg},t} 
    &= \phi_{\mathrm{hdg}}\!\left(l_{\mathrm{hdg},t},\,\tilde{l}_{\mathrm{hdg}}\right) \\
    &= w_{\mathrm{hdg}} \cdot g_t\cdot
    \mathrm{ReLU}\!\left(
    \max(\underline{d} - \hat{d}_{i^\ast},\;
    \hat{d}_{i^\ast} - \overline{d})
    \right)^{2},
\end{aligned}
\label{eq:phi_hdg}
\end{equation}
where $w_{\mathrm{hdg}}$ is a scaling factor and $g_t$ is a proximity-aware gate function defined as
\begin{equation}
    g_t
    = \mathrm{sigmoid}\!\left(
    \frac{10-\|\mathbf{p}_{i^\ast,t}-\mathbf{p}_{\mathrm{ego},t}\|}{3}
    \right),
\end{equation}
which approaches zero when the adversarial vehicle is far from the ego vehicle and gradually increases toward one as it gets closer. Therefore, the constraint on $h_{\mathrm{hdg},t}$ penalizes the squared deviation of the predicted relative heading indicator $\hat{d}_{i^\ast}$ from the admissible interval $[\underline{d},\overline{d}]$ associated with the target collision type $m^*$. The penalty is modulated by the gate function, so that the heading alignment constraint becomes active primarily when the adversarial vehicle is sufficiently close to the ego vehicle.

\subsection{Severity Constraint}
At timestep $t$, we further define $l_{\mathrm{svt},t}$ as the predicted closing speed at the target impact timestep $t+T_{\mathrm{col}}$:
\begin{equation}
    l_{\mathrm{svt},t}
    =
    \hat{\mathbf{r}}(m^\ast)^\top
    \left(
    \dot{\hat{\mathbf{p}}}_{i^\ast}
    -
    \dot{\hat{\mathbf{p}}}_{\mathrm{ego}}
    \right),
\end{equation}
which projects the relative velocity between the adversarial agent $i^\ast$ and ego vehicle at the target impact timestep, $\dot{\hat{\mathbf{p}}}_{i^\ast}-\dot{\hat{\mathbf{p}}}_{\mathrm{ego}}$, onto $\hat{\mathbf{r}}(m^\ast)$, the unit direction corresponding to the target collision type $m^\ast$. 

We define the collision-relevant unit direction as:
\begin{equation}
\hat{\mathbf{r}}(m^\ast)
=
\begin{cases}
\hat{\mathbf n}_{\mathrm{ego}}, & \text{Rear-end}\\
-\hat{\mathbf n}_{\mathrm{ego}}, & \text{Head-on}\\
\hat{\mathbf n}_{\mathrm{ego}}^{\perp}, & \text{Side}\\
\hat{\mathbf n}_{i^*}, & \text{Cut-in}
\end{cases}
\end{equation}
where $\hat{\mathbf n}_{\mathrm{ego}}^{\perp}$ denotes the lateral axis of the ego vehicle at $t+T_\mathrm{col}$:
\begin{equation}
\hat{\mathbf n}_{\mathrm{ego}}^{\perp}
=
\begin{cases} 
[\sin\hat{\vartheta}_{\mathrm{ego}},\; -\cos\hat{\vartheta}_{\mathrm{ego}}]^\top, & \text{if } i^* \text{ approaches from the left.} \\ 
[-\sin\hat{\vartheta}_{\mathrm{ego}},\; \cos\hat{\vartheta}_{\mathrm{ego}}]^\top, & \text{if } i^* \text{ approaches from the right.}
\end{cases}
\end{equation}

$\tilde{l}_\mathrm{svt}$ denotes the lower bound required for the closing speed in a collision of type $m^\ast$, as specified in Table \ref{tab:constraints}. Then, a dissimilarity function measures the closing speed residual:
\begin{equation}
    h_{\mathrm{svt},t} = \phi_{\mathrm{svt}}\!\left(l_{\mathrm{svt}},\,\tilde{l}_{\mathrm{svt}}\right)
    = w_{\mathrm{svt}} \cdot \mathrm{ReLU}\!\left(\tilde{l}_{\mathrm{svt}} - l_{\mathrm{svt}}\right),
    \label{eq:phi_svt}
\end{equation}
where $w_{\mathrm{svt}}$ is a scaling factor. 
$h_{\mathrm{svt},t}$ measures the shortfall of impact severity.
Therefore, the corresponding constraint penalizes cases where the predicted closing speed $l_{\mathrm{svt},t}$ falls below the required lower bound $\tilde{l}_{\mathrm{svt}}$ associated with the target collision type.
The penalty increases linearly with the amount of closing speed shortfall and becomes zero once the closing speed meets or exceeds the required threshold.

\begin{table}[h]
\centering
\caption{Collision-type-specific constraint thresholds}
\label{tab:constraints}

\newcolumntype{L}[1]{>{\raggedright\arraybackslash}p{#1}}
\newcolumntype{R}[1]{>{\raggedleft\arraybackslash}p{#1}}
\newcolumntype{C}[1]{>{\centering\arraybackslash}p{#1}}

\begin{tabular}{L{2cm}|C{2cm}|C{4cm}|C{3cm}}
\toprule
$m^*$ 
& $\tilde{l}_{\mathrm{cnt}}$
& $\tilde{l}_{\mathrm{hdg}}$
& $\tilde{l}_{\mathrm{svt}}
$ (m/s)\\
\midrule

Rear-end
& $-10^{-6}$
& $[0.95,\,1.00]$
& 2 \\

Side
& $-10^{-6}$
& $[-0.17,\,0.17]$
& 1 \\

Cut-in
& $-10^{-6}$
& $[0.71,\,0.97]$
& 4 \\

Head-on
& $-10^{-6}$
& $[-1.00,\,-0.95]$
& 7 \\

\bottomrule
\end{tabular}
\end{table}

\section{Gauss--Newton Projection Details}
\label{sec:projection_SM}

 This section details two components abstracted in the main text: (i) the dynamic time-to-collision $T_{\mathrm{col}}$ used in the residual $\mathbf{h}_t$, and (ii) the damped Gauss--Newton projection that implements $\Pi_{\mathcal{C}}$.

\subsection{Dynamic time-to-collision}
\label{sec:dynamic_tcol}

The target collision horizon $T_{\mathrm{col}}$ is updated at the same frequency as the closed-loop re-planning. At the beginning of each re-planning step, we recompute $T_{\mathrm{col}}$ using the states of the ego vehicle and the adversarial agent observed at the current closed-loop timestep $t$.

Specifically, we estimate the time to closest approach under a constant-velocity assumption and convert it into a discrete planning-step index:
\begin{equation}
T_{\mathrm{col}}
=
\mathrm{clip}\!\left(
\left\lfloor
\frac{1}{dt}
\frac{-\Delta \mathbf{v}_{t}^{\top}\Delta\mathbf{p}_{t}}
{\|\Delta\mathbf{v}_{t}\|^{2}}
\right\rfloor,
T_{\min},
T_{\max}
\right),
\label{eq:dynamic_tcol}
\end{equation}
where $dt$ is the simulation step time, $\Delta\mathbf{p}_{t}(=\mathbf{p}_{\mathrm{ego},t}-\mathbf{p}_{{i^*},t})$ and $\Delta\mathbf{v}_{t}(=\mathbf{v}_{\mathrm{ego},t}-\mathbf{v}_{{i^*},t})$ denote the relative position and relative velocity between the adversarial agent and the ego vehicle, respectively. The clipping range $[T_{\min},T_{\max}]=[5,10]$ constrains the target collision horizon to a practical range: a horizon that is too short may lead to an infeasible or collapsed constraint window, whereas a horizon that is too long may delay the induced collision. 

When $\|\Delta\mathbf{v}_{t}\|^{2}$ is close to zero or when $\Delta \mathbf{v}_{t}^{\top}\Delta\mathbf{p}_{t}\geq 0$, the estimated closest-approach time is undefined, unstable, or non-positive. In these cases, we set $T_{\mathrm{col}}=T_{\max}$. Updating $T_{\mathrm{col}}$ at every re-planning step keeps the constraint window aligned with the evolving ego--adversary geometry during closed-loop simulation.


\subsection{Damped Gauss--Newton projection}
\label{sec:gn_projection}
Given the unprojected Euler step $\hat{\mathbf{a}}_{i^*}(\lambda_{k+1})$, the damped Gauss--Newton projection computes a locally minimal correction that moves the action sequence toward the collision-constraint set. Formally, the projection can be written as
\begin{equation}
\Pi_{\mathcal{C}}(\hat{\mathbf{a}}_{i^*}(\lambda_{k+1}))
\;=\;
\arg\min_{\mathbf{a}_{\mathrm{proj}}\in\mathcal{C}}\;\|\mathbf{a}_{\mathrm{proj}}-\hat{\mathbf{a}}_{i^*}(\lambda_{k+1})\|^2.
\label{eq:gn_proj_obj}
\end{equation}
Because the residual $\mathbf{h}_t$ is evaluated using the adversarial-agent state at the target future step $t+T_{\mathrm{col}}$, only the actions up to that step can influence the residual through the forward dynamics. Under the causal discrete-time rollout, the state at $t+T_{\mathrm{col}}$ depends on the actions from $t$ to $t+T_{\mathrm{col}}-1$, but not on the later actions $\mathbf{a}_{i^*,t+T_{\mathrm{col}}:t+T-1}$. Therefore, the projection only updates the first $T_{\mathrm{col}}$ action steps. Within this section, we denote this action slice as 
$\mathbf{a}\in\mathbb{R}^{T_{\mathrm{col}}\times 2}$ and use 
$\mathcal{F}(\mathbf{s}_{i^*,t},\mathbf{a})$ to denote the corresponding 
$T_{\mathrm{col}}$-step rollout.

The Jacobian
\begin{equation}
\mathbf{J} \;=\; \frac{\partial \mathbf{h}\!\left(\mathcal{F}(\mathbf{s}_{i^*,t},\mathbf{a})\right)}{\partial \mathbf{a}}\;\in\;\mathbb{R}^{n_h\times 2T_{\mathrm{col}}},
\label{eq:jacobian_autograd}
\end{equation}
is obtained by reverse-mode automatic differentiation through the differentiable forward-dynamics rollout $\mathcal{F}$, where $n_h=3$ is the number of residual components in $\mathbf{h}_t$. We then take a damped Gauss--Newton Projection step:
\begin{equation}
\mathbf{a} \;\leftarrow\; \mathbf{a} - \alpha\,\mathbf{J}^{\top}\!\left(\mathbf{J}\mathbf{J}^{\top} + \gamma\,\mathbf{I}\right)^{-1}\mathbf{h}\!\left(\mathcal{F}(\mathbf{s}_{i^*,t},\mathbf{a})\right),
\label{eq:gn_update}
\end{equation}
where the step size is set to be $\alpha=0.8$ and the Levenberg--Marquardt damping is set to $\gamma=10^{-4}$. The damping term improves the condition of $(\mathbf{J}\mathbf{J}^{\top}+\gamma\mathbf{I})$ when the residual rows are nearly colinear, and $\alpha<1$ mitigates overshooting that may occur with a full GN update under highly nonlinear constraints. Applying $\Pi_{\mathcal{C}}$ at every ODE step (combined with the OT reverse update) progressively drives the trajectory toward $\mathcal{C}$ as the flow time $\lambda$ approaches $1$. The full procedure is summarized in Algorithm \ref{alg:projection}.

\begin{algorithm}[h]
\caption{Damped Gauss--Newton Projection $\Pi_{\mathcal{C}}$}
\label{alg:projection}
\begin{algorithmic}[1]
\State \textbf{Input:}
    Unprojected Euler step $\hat{\mathbf{a}}_{i^*}(\lambda_{k+1})$,
    adversarial initial state $\mathbf{s}_{i^*,t}$,
    ego trajectory $\boldsymbol{\tau}_{\mathrm{ego},t:t+T}$,
    target collision type $m^*$,
    dynamic time-to-collision $T_{\mathrm{col}}$
\State \textbf{Output:} Projected action sequence $\mathbf{a}_{\mathrm{proj}}$
\State $\mathbf{a} \gets \hat{\mathbf{a}}_{i^*}(\lambda_{k+1})_{[\,0:T_{\mathrm{col}}-1\,]}$ \Comment{Slice to first $T_{\mathrm{col}}$ steps}
\State $\boldsymbol{\tau} \gets \mathcal{F}(\mathbf{s}_{i^*,t},\, \mathbf{a})$ \Comment{Forward-dynamics rollout}
\State $\mathbf{h} \gets \phi_j(l_{j,t}, \tilde{l}_j),\quad \text{for } j \in \{\mathrm{cnt}, \mathrm{hdg}, \mathrm{svt}\}$ \Comment Evaluation of residuals
\State $\mathbf{J} \gets \partial \mathbf{h} / \partial \mathbf{a}$ \Comment Via reverse-mode automatic differentiation through $\mathcal{F}$
\State $\mathbf{a} \gets \mathbf{a} - \alpha\,\mathbf{J}^{\top}\!\left(\mathbf{J}\mathbf{J}^{\top} + \gamma\mathbf{I}\right)^{-1}\mathbf{h}$ \Comment{Damped GN step}
\State Clamp $\mathbf{a}$ to valid acceleration and yaw-rate bounds
\State $\mathbf{a}_{\mathrm{proj}} \gets$ concatenate $\mathbf{a}$ with the unchanged tail $\hat{\mathbf{a}}_{i^*}(\lambda_{k+1})_{[\,T_{\mathrm{col}}:T\,]}$
\State \Return $\mathbf{a}_{\mathrm{proj}}$
\end{algorithmic}
\end{algorithm}

\section{Details on Training and Inference Settings}
 In closed-loop simulation, each agent's action selection is conditioned on its context $\mathbf{c}_{i,t}$, which encapsulates:
(i) an agent-centric rasterized map $\mathbf{I}_i \in \mathbb{R}^{H \times W \times C}$ encoding road geometry and lane structure; and
(ii) historical states of neighboring vehicles
$\mathbf{s}_{t-T_{\text{hist}}:t} \in \mathbb{R}^{(N-1) \times T_{\text{hist}} \times 4}$ over the past $T_{\text{hist}}=10$ timesteps. 
The CCFM framework is trained on 2 $\times$ H200 GPUs using AdamW optimizer for 30,000 steps. The initial learning rate is set to $1\times10^{-5}$, with a batch size of 1,024 per GPU. CCFM is trained to predict 32 future frames conditioned on historical frames $T_{\text{hist}}=10$, with a step-time 0.1 s. During closed-loop simulation, all experiments are conducted on a single L40S GPU.

Meanwhile, the details of inference are as follows:
\begin{itemize}
    \item \textit{Heuristic Collision Selector.}  In closed-loop simulation, we dynamically select the collision event type and the adversarial vehicle pair at a frequency of 2~Hz, consistent with the ego re-planning frequency. The three scoring terms are assigned equal weights.
    \item \textit{Sampling Strategy.} We use an Euler solver with 20 flow matching steps (i.e., $K=20$). To preserve driving realism and reduce unnecessary collisions with other reactive agents, we incorporate realism guidance and generate 20 candidate future trajectories. Among them, we select the trajectory with the lowest cost, following the strategy in SAFE-SIM \cite{safe-sim}.
\end{itemize}

\section{Evaluation Metrics Definition}

We define the metrics  for evaluating CCFM and its benchmarks. All metrics are calculated from the states of the adversarial and ego vehicles at the actual collision time $t_\mathrm{col}$.

\subsection{Scenario Criticality}
Collision Rate (CR) measures the fraction of closed-loop simulations in which a collision occurs between the ego and adversarial vehicles, computed as:
\begin{equation}
    \mathrm{CR}=\frac{1}{M}
\sum_{q=1}^{M}
1(\text{collision occurs in scenario } q),
\end{equation}
where $M$ is the total number of simulated scenarios.

Collision severity (MS) is another indicator for scenario criticality. To characterize the severity of the collision, we report the relative speed at impact. For a collided pair of ego and adversarial vehicles, the magnitude of relative speed at impact is defined as:
\begin{equation}
\mathrm{MS}
=\frac{1}{M_{\mathrm{col}}}
\sum_{n=1}^{M_{\mathrm{col}}}
\left\|
\dot{\mathbf{p}}_{\mathrm{ego},t_\mathrm{col}}^n - \dot{\mathbf{p}}_{i^*,t_\mathrm{col}}^n
\right\|_2 (\text{across collided scenario } n).
\end{equation}
We further define the Criticality Score (CS) to jointly summarize collision rate and collision severity. 
Since higher CR and higher MS indicate more safety-critical scenarios, both metrics are first normalized by the maximum value among the compared methods under the same experimental setting:
\begin{equation}
\widehat{\mathrm{CR}\cdot \mathrm{TM}} =
\frac{\mathrm{CR}\cdot \mathrm{TM}}{\max(\mathrm{CR}\cdot \mathrm{TM})},
\quad
\widehat{\mathrm{MS}} =
\frac{\mathrm{MS}}{\max(\mathrm{MS})},
\end{equation}
where  $\mathrm{TM}$ is Type Match defined in Eq. \ref{TM}, and $\mathrm{TM}=1$ if Type-Match rate is not available. 
The Criticality Score is then computed as
\begin{equation}
\mathrm{CS}
=
\frac{1}{2}
\left(
\widehat{\mathrm{CR} \cdot \mathrm{TM}}
+
\widehat{\mathrm{MS}}
\right),
\end{equation}
A higher CS indicates stronger overall scenario criticality in terms of both collision occurrence and impact severity. 

\subsection{Behavioral Realism}

The Off-Road (OR) rate measures the proportion of scenarios in which reactive agents leave the drivable area, computed as:
\begin{equation}
\mathrm{OR}
=
\frac{1}{M}
\sum_{q=1}^{M}
1(\text{reactive agent that goes off-road in scenario } q
).
\end{equation}
Correspondingly, the adversarial Off-Road rate (ADV.OR) evaluates the proportion that the adversarial agent goes off-road among scenarios where a collision occurs. It is computed as:
\begin{equation}
\mathrm{ADV.OR}
=
\frac{1}{M_{\mathrm{col}}}
\sum_{n=1}^ {M_{\mathrm{col}}} 1(\text{adversarial agent goes off-road in collided scenario } n),
\end{equation}
where $M_{\mathrm{col}}$ denotes the number of generated scenarios in which a collision occurs between ego agent and adversarial agent.

To quantify motion realism, we compare the distributions of longitudinal acceleration ($\dot{v}_\mathrm{lon}$), lateral acceleration ($\dot{v}_\mathrm{lat}$), and jerk ($\ddot{v}$) between simulated and real trajectories for reactive agents. For each attribute, we construct normalized histograms and compute the Wasserstein distance $\mathcal{W}_1(\cdot)$ between the simulated and real distributions. The realism deviation is defined as the average Wasserstein distance across the three motion attributes:
\begin{equation}
\mathrm{RM}
=
\frac{1}{3}
\sum_{v \in \{\dot{v}_\mathrm{lon},\dot{v}_\mathrm{lat},\ddot{v}\}}
\mathcal{W}_1(p_v,q_v),
\end{equation}
where \(p_v\) and \(q_v\) denote the normalized histograms of the simulated and real data, respectively, for motion attribute $v\in \{\dot{v}_\mathrm{lon},\dot{v}_\mathrm{lat},\ddot{v}\}$.
We further define the Realism Score (RS) to jointly summarize off-road safety and motion realism across scenarios. Since lower OR and lower RM indicate more realistic behaviors, both metrics are converted into positive scores by comparing them with the minimum value among the compared methods under the same experimental setting:
\begin{equation}
\widehat{\mathrm{OR}}
=
\frac{\min(\mathrm{OR})}{\mathrm{OR}},
\quad
\widehat{\mathrm{RM}}
=
\frac{\min(\mathrm{RM})}{\mathrm{RM}}.
\end{equation}
The Realism Score is then computed as
\begin{equation}
\mathrm{RS}
=
\frac{1}{2}
\left(
\widehat{\mathrm{OR}}
+
\widehat{\mathrm{RM}}
\right).
\end{equation}
A higher RS indicates better overall behavioral realism, with fewer off-road violations and smaller deviation from ground truth motion distributions.

We further report displacement-based realism metrics, including Average Displacement Error (ADE) and Final Displacement Error (FDE). Let $\hat{\mathbf{p}}_{i,t}^q$ and $\mathbf{p}_{i,t}^q$ denote the generated and ground-truth positions of agent $i$ in scenario $q$ at future timestep $t$, respectively, and let $N_q$ denote the number of agents in the $q$-th scenario. ADE and FDE are computed over all generated trajectories as: 

\begin{equation} \mathrm{ADE} = \frac{1}{T \sum_{q=1}^{M} N_q} \sum_{q=1}^{M} \sum_{i=1}^{N_q} \sum_{t=1}^{T} \left\| \hat{\mathbf{p}}_{i,t}^q - \mathbf{p}_{i,t}^q \right\|_2, 
\end{equation} 
\begin{equation} \mathrm{FDE} = \frac{1}{\sum_{q=1}^{M} N_q} \sum_{q=1}^{M} \sum_{i=1}^{N_q} \left\| \hat{\mathbf{p}}_{i,T}^q - \mathbf{p}_{i,T}^q \right\|_2. 
\end{equation} 

Correspondingly, ADV.ADE and ADV.FDE evaluate the displacement errors of the adversarial agent among scenarios where a collision occurs. Let $i_n^{*}$ and $t_{\mathrm{col},n}$ denote the adversarial agent and the actual collision timestep in the $n$-th collided scenario, respectively. They are computed as: 
\begin{equation} \mathrm{ADV.ADE} = \frac{1}{\sum_{n=1}^{M_\mathrm{col}} t_{\mathrm{col},n}} \sum_{n=1}^{M_{\mathrm{col}}} \sum_{t=1}^{t_{\mathrm{col},n}} \left\| \hat{\mathbf{p}}_{i_n^{*},t}^n - \mathbf{p}_{i_n^{*},t}^n \right\|_2, 
\end{equation} 
\begin{equation} \mathrm{ADV.FDE} = \frac{1}{M_{\mathrm{col}}} \sum_{n=1}^{M_\mathrm{col}} \left\| \hat{\mathbf{p}}_{i_n^{*},t_{\mathrm{col},n}}^n - \mathbf{p}_{i_n^{*},t_{\mathrm{col},n}}^n \right\|_2,
\end{equation}

\subsection{Scenario Controllability and Diversity}
To assess the controllability of the actual collision type with respect to the target collision type, we formulate the collision Type-Match (TM) rate as:
\begin{equation}
\mathrm{TM}
=
    \frac{1}{M_{\mathrm{col}}}
\sum_{n=1}^{M_{\mathrm{col}}}
1(m_n^{\mathrm{act}} = m_n^{\mathrm{tar}}),
\label{TM}
\end{equation}
where $m_n^{\mathrm{act}}$ and $m_n^{\mathrm{tar}}$ denote the actual and target collision types for the $n$-th collided scenario, respectively, and $M_{\mathrm{col}}$ is the number of samples that result in a collision between the ego and adversarial vehicles.

To evaluate the diversity of collision events, we further report the contact region on the ego vehicle at impact. 
For each collision scenario, the contact location is categorized into front, rear, or side according to the geometry of the first collision frame. 
The corresponding rate for region $r$ is defined as
\begin{equation}
\mathrm{Region}_{r}
=
\frac{1}{M_{\mathrm{col}}}
\sum_{n=1}^{M_{\mathrm{col}}}
\mathbf{1}(\mathrm{reg}_n = r),
\end{equation}
where $\mathrm{reg}_n$ denotes the contact-region label of the $n$-th collision scenario, and $r\in \mathcal{R}=\{\mathrm{front}, \mathrm{rear}, \mathrm{side}\}$ denotes a contact region category. 
The normalized entropy of the contact-region distribution, denoted as $\mathrm{EN}$, is computed as
\begin{equation}
\mathrm{EN}
=
-\frac{1}{\log |\mathcal{R}|}
\sum_{r \in \mathcal{R}}
\mathrm{Region}_{r}
\log
\left(
\mathrm{Region}_{r}
\right),
\end{equation}
where we use the convention $0\log 0=0$. EN takes values from the range [0, 1]: it is zero when all collisions occur in one contact region, and one if collisions are uniformly distributed across impact region categories. Therefore, a higher $\mathrm{EN}$ indicates a more diverse and balanced distribution of collision contact regions.

Additionally, we report the standard deviation of the relative speed (SD) and relative heading angle (HD) between the ego and adversarial vehicles across all scenarios of collision. The standard deviation of relative speed SD is defined as:
\begin{equation}
\mathrm{SD}
=\frac{1}{M_{\mathrm{col}}}
\sum_{n=1}^{M_{\mathrm{col}}}
\operatorname{Std}
\left(
\left\|
\dot{\mathbf{p}}_{\mathrm{ego},t_\mathrm{col},n}^{n}
-
\dot{\mathbf{p}}_{i^*,t_\mathrm{col},n}^{n}
\right\|
\right).
\end{equation}
where $\mathbf{p}_{\mathrm{ego},t_{\mathrm{col},n}}^{n}$ and $\mathbf{p}_{i^*,t_{\mathrm{col},n}}^{n}$ denote the position vectors of the ego vehicle and the adversarial agent $i^\ast$ at the collision time $t_{\mathrm{col},n}$ in the $n$-th collision event.

The standard deviation of the relative heading angle HD is defined as:
\begin{equation}
\mathrm{HD}
=\frac{1}{M_{\mathrm{col}}}
\sum_{n=1}^{M_{\mathrm{col}}}
\operatorname{Std}
\left(
\arccos\left(
\mathbf{n}_{\mathrm{ego},t_{\mathrm{col},n}}^{n}
\cdot
\mathbf{n}_{i^*,t_{\mathrm{col},n}}^{n}
\right)
\right),
\end{equation}
where $\mathbf{n}_{\mathrm{ego},t_{\mathrm{col},n}}^{n}$ and $\mathbf{n}_{i^*,t_{\mathrm{col},n}}^{n}$ denote the unit heading vectors of the ego vehicle and adversarial vehicle $i^*$ at the collision time $t_{\mathrm{col},n}$ for the $n$-th collision event.

Based on these metrics, we define the Diversity Score (DS) to jointly measure the diversity of contact-region distribution, relative speed, and relative heading angle. 
Since EN is already normalized to $[0,1]$, we only normalize SD and HD by the maximum value among the compared methods under the same experimental setting:
\begin{equation}
\widehat{\mathrm{SD}}
=
\frac{\mathrm{SD}}{\max(\mathrm{SD})},
\quad
\widehat{\mathrm{HD}}
=
\frac{\mathrm{HD}}{\max(\mathrm{HD})}.
\end{equation}
The Diversity Score is then computed as
\begin{equation}
\mathrm{DS}
=
\frac{1}{3}
\left(
\mathrm{EN}
+
\widehat{\mathrm{SD}}
+
\widehat{\mathrm{HD}}
\right).
\end{equation}
A higher DS indicates more diverse collision outcomes in terms of contact location, impact-speed variation, and relative heading geometry.

\subsection{Overall Composite Weighted Score}
To provide an overall comparison across scenario criticality, diversity, and behavioral realism, we define the Composite Weighted Score (CWS) by combining the Criticality Score (CS), Diversity Score (DS), and Realism Score (RS):

\begin{equation}
\mathrm{CWS}=\frac{1}{3}\left(\mathrm{CS}+\mathrm{DS}+\mathrm{RS}\right).\end{equation}
A higher CWS indicates better overall performance, reflecting a more comprehensive trade-off among safety-critical scenario generation, collision diversity, and behavioral realism.
\section{Details on Supplementary Experiments}

In this section, we provide additional experiments and analyses to further evaluate the controllability and robustness of CCFM, including quantitative collision-type controllability results and failure cases and infeasible constraints analysis.

\subsection{Quantitative Controllability Analysis}
The results in Table \ref{tab:tm_confusion} further demonstrate the collision-type controllability of CCFM. The diagonal entries remain consistently high across all target collision types, ranging from 83.4\% to 91.7\%, indicating that the generated collisions largely match the target types selected by HCS. The majority of off-diagonal cases occur between geometrically related collision types, such as rear-end collisions being generated as cut-in collisions and side collisions being generated as cut-in collisions. These discrepancies are expected in closed-loop interaction, where the relative pose between the ego vehicle and the adversarial agent can evolve over time and consequently shift the apparent collision type. Overall, the confusion matrix shows that CCFM provides reliable semantic control over collision outcomes rather than merely increasing the collision rate.

\begin{table}[t]
\caption{Quantitative collision-type controllability analysis. The table reports the Type-Match (TM) confusion matrix between the target collision type selected by HCS and the actual collision type observed in closed-loop simulation. Diagonal entries indicate successful type-level control, and off-diagonal entries indicate mismatched collision outcomes.}
\label{tab:tm_confusion}
\centering
\setlength{\tabcolsep}{5pt}
\begin{tabular}{l|cccc}
\toprule
\diagbox{Target}{Actual}
& \textbf{Rear-end}
& \textbf{Head-on}
& \textbf{Side}
& \textbf{Cut-in} \\
\midrule
\textbf{Rear-end}
& \textbf{\textcolor{brown}{83.4\%}} & 1.2\% & 3.6\% & 11.2\% \\

\textbf{Head-on}  
& 0.0\% & \textbf{\textcolor{brown}{91.7\%}} & 8.3\% & 0.0\% \\

\textbf{Side}     
& 0.0\% & 0.0\% & \textbf{\textcolor{brown}{90.0\%}} & 10.0\% \\

\textbf{Cut-in}   
& 7.7\% & 0.0\% & 7.7\% & \textbf{\textcolor{brown}{84.6\%}} \\
\bottomrule
\end{tabular}
\end{table}

\begin{table}[htbp]
\caption{Per-collision-type analysis of failure cases and infeasible-constraint analysis. We report the success and failure rates for each HCS-selected target collision type, together with the cumulative infeasibility ratios of the contact-point, relative-heading, and severity constraints in collided scenarios, denoted as $\mathrm{Inf}_{\mathrm{cnt}}$, $\mathrm{Inf}_{\mathrm{hdg}}$, and $\mathrm{Inf}_{\mathrm{svt}}$, respectively.}
\label{tab:failure}
\centering
\setlength{\tabcolsep}{6pt}
\begin{tabular}{l|ccccc}
\toprule
\textbf{HCS Target} 
& \textbf{Success}(\%) 
&\textbf{Failure}(\%)
&$\mathrm{Inf}_\mathrm{cnt}$(\%)
&$\mathrm{Inf}_\mathrm{hdg}$(\%)
&$\mathrm{Inf}_\mathrm{svt}$ (\%)\\
\midrule
Rear-end & 83.4 & 16.6 & 28.77 & 0.85& 2.97\\
Head-on  & 91.7 & 8.3 & 32.86 & 1.43&\textbf{\textcolor{brown}{39.29}} \\
Side     & 90.0 & 10.0 & \textbf{\textcolor{brown}{72.86}} &7.14 & 11.43\\
Cut-in   & 84.6 & 15.4 & \textbf{\textcolor{brown}{55.64}} & 6.39 &19.17\\
\bottomrule
\end{tabular}
\end{table}

\subsection{Failure Cases and Infeasible-constraint Analysis}

Table \ref{tab:failure} also provides the cumulative infeasibility ratios of the collision constraints $(\mathrm{Inf}_{\mathrm{cnt/hdg/svt}})$ in collided scenarios, evaluated against kinematic feasibility bounds at $T_{\mathrm{col}}$. Among the constraint components, side and cut-in scenarios are primarily limited by contact-point feasibility, indicating that lateral reachability and precise contact alignment are more challenging for these interaction patterns. In contrast, head-on scenarios are mainly constrained by severity feasibility, reflecting the difficulty of satisfying the required relative-speed condition under closed-loop simulation. Nevertheless, through successive re-planning, the adversary can progressively correct infeasible intermediate targets, continue the intended maneuver, and ultimately induce a collision with the ego vehicle.




%
%

\end{document}